%% file: main.tex
\documentclass{article} 
\usepackage{iclr2024_conference,times}

\input{math_commands.tex}

\usepackage{hyperref}
\usepackage{url}
\usepackage{amsthm}

\usepackage{mathabx}
\usepackage{comment}
\usepackage{graphicx}
\usepackage{placeins}
\usepackage{floatrow}
\usepackage[super]{nth}
\usepackage[font=footnotesize]{caption}

\title{Identifying Interpretable Visual Features in \\Artificial and Biological Neural Systems}

\author{David Klindt\\
Stanford University\\
\texttt{klindt.david@gmail.com} \\
\And
Sophia Sanborn\\
UC Santa Barbara\\
\texttt{sanborn@ucsb.edu} \\
\And
Francisco Acosta\\
UC Santa Barbara\\
\texttt{facosta@ucsb.edu} \\
\And
Fr\'ed\'eric Poitevin\\
SLAC National Accelerator Laboratory\\
\texttt{frederic.poitevin@stanford.edu} \\
\And
Nina Miolane\\
UC Santa Barbara\\
\texttt{ninamiolane@ucsb.edu}
}

\iclrfinalcopy 
\begin{document}

\maketitle

\begin{abstract}
\input{sections/0-abstract}
\end{abstract}

\input{sections/1-introduction}
\input{sections/2-methods}

\input{sections/3-results}

\input{sections/4-discussion}

\input{sections/6-acknowledgements}

\bibliography{references}
\bibliographystyle{iclr2024_conference}

\input{sections/5-appendix}

\end{document}

%% file: math_commands.tex
\usepackage{amsmath,amsfonts,bm}

\def\eqref#1{equation~\ref{#1}}

\def\1{\bm{1}}

\DeclareMathAlphabet{\mathsfit}{\encodingdefault}{\sfdefault}{m}{sl}
\SetMathAlphabet{\mathsfit}{bold}{\encodingdefault}{\sfdefault}{bx}{n}



%% file: sections/0-abstract.tex
Single neurons in neural networks are often interpretable in that they represent individual, intuitively meaningful features. However, many neurons exhibit \textit{mixed selectivity}, i.e., they represent multiple unrelated features. A recent hypothesis proposes that features in deep networks may be represented in \textit{superposition}, i.e., on non-orthogonal axes by multiple neurons, since the number of possible interpretable features in natural data is generally larger than the number of neurons in a given network. Accordingly, we should be able to find meaningful directions in activation space that are not aligned with individual neurons. Here, we propose (1) an automated method for quantifying visual interpretability that is validated against a large database of human psychophysics judgments of neuron interpretability, and (2) an approach for finding meaningful directions in network activation space. We leverage these methods to discover directions in convolutional neural networks that are more intuitively meaningful than individual neurons, as we confirm and investigate in a series of analyses. Moreover, we apply the same method to three recent datasets of visual neural responses in the brain and find that our conclusions largely transfer to real neural data, suggesting that superposition might be deployed by the brain. This also provides a link with disentanglement and raises fundamental questions about robust, efficient and factorized representations in both artificial and biological neural systems.

%% file: sections/1-introduction.tex
\section{Introduction}
\vspace{-0.2cm}
\begin{figure}[h!]
    \floatbox[{\capbeside\thisfloatsetup{capbesideposition={right,top},capbesidewidth=5.1cm}}]{figure}[\FBwidth]
    {\caption{
        \textbf{Conceptual Overview.}
        \textbf{A}) A representation of two neurons' activations for different images. The highlights indicate maximally exciting images (MEIs) for each neuron.
        \textbf{B}) There exist directions in feature space that are more interpretable.
        }
        \label{fig:project_overview}}
    {\includegraphics[width=8.3cm]{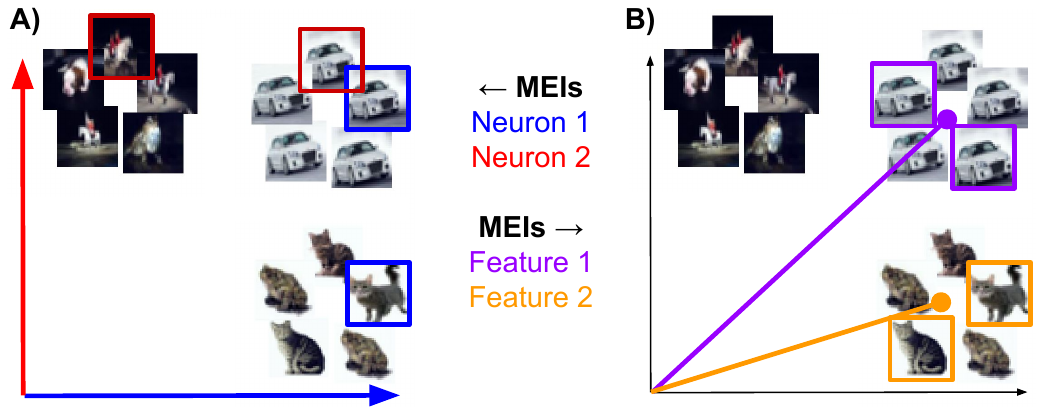}}
\end{figure}
\vspace{-.2cm}

One of the oldest ideas in neuroscience is Cajal's \textit{single neuron doctrine} \citep{finger2001origins} and its application to perception \citep{barlow1972single}, i.e., the hypothesis that individual sensory neurons encode individually meaningful \textit{features}.\footnote{In this work, we adopt a pragmatic definition of \textit{feature} based on human discernability, measured through psychophysics experiments (see below). For an attempt at a more formal definition see \cite{elhage2022toy}.} The idea dates back to the early 1950s, when researchers began to find evidence of neurons that reliably and selectively fire in response to particular stimuli, such as dots on a contrasting background \citep{barlow1953summation} and lines of particular orientation and width \citep{hubel1959receptive}. These findings gave rise to the \textit{standard model} of the ventral visual stream as a process of hierarchical feature extraction and pooling \citep{hubel1968receptive, gross1972visual,  riesenhuber1999hierarchical, quiroga2005invariant}. Neurons in the early stages extract simple features, such as oriented lines, while neurons at later stages combine simple features to construct more complex composite features. In the highest stages, complex features are combined to yield representations of entire objects encoded by single neurons\textemdash the shape of a hand, or the face of a friend. 
Notwithstanding a shift in focus towards population codes \citep{averbeck2006neural,stanley2013reading,hebb2005organization,gao2015simplicity,jacobs2009ruling,ebitz2021population}, this model has remained a dominant paradigm in sensory neuroscience for the last seven decades and ultimately inspired \citep{hassabis2017neuroscience,zador2023catalyzing} the development of convolutional neural networks (CNNs) \citep{fukushima1980neocognitron,lecun1989backpropagation} \citep[but see][]{gershman2023have,poggio2017and}.

Mechanistic interpretability research aims to uncover the coding properties of neurons within artificial neural networks. \textit{Feature visualization} \citep{nguyen2019understanding}\textemdash i.e. the single-unit electrophysiology of artificial neural networks\textemdash has revealed remarkable consistencies between neurons in image models and neurons in the visual cortex: neurons with center-surround receptive fields, color-contrast detectors, and oriented edge detectors that combine to form curve detectors in higher layers, for example \citep{olah2020overview,willeke2023deep}.
However, the study of individual neurons, both \textit{in vitro} and \textit{in silico}, faces two major problems. First is the inherent subjectivity of ``interpretability,'' which generally necessitates the hand-inspection of neuron response properties. Second is the ubiquitous existence of hard-to-interpret units with mixed selectivity \citep{fusi2016neurons, olah2020overview}.\footnote{One might wonder why evolution or gradient descent would be so kind as to make any neurons interpretable.
Anecdotally, researchers have explained this as a result of the use of pointwise nonlinearities in deep networks.
We provide a more formal argument for this explanation in App.~\ref{sec:privileged-basis}.} We address both problems in this work by (1) defining a quantitative, automated measure of interpretability for vision models that does not rely on human inspection, and (2) demonstrating a simple approach for finding meaningful directions in activity space. 

A recent paper by \cite{zimmermann2023scale} has taken a similar approach, using human perceptual judgments in large-scale psychophysics experiments to quantify the interpretability of neurons within deep image models \citep{zimmermann2021well,borowski2020exemplary}. 
We automate this pipeline by replacing human judgments of perceptual similarity with a similarity metric grounded in deep image model representations \citep{zhang2018unreasonable}, and validate the approach against the large scale human data from \cite{zimmermann2023scale}. 
Thus, in line with recent work that uses \textit{language models to interpret language models} \citep{bills2023language}, we use \textit{image models to interpret image models}.
We then leverage this automated index of interpretability to test whether non-axis aligned directions in the neural activation space of CNNs trained on real data may be more interpretable than the individual units\textemdash a test of the recently stated \textit{superposition hypothesis} \citep{elhage2022toy}. 
 
Through a suite of experiments and analyses, we find evidence consistent with the hypothesis that neurons in both deep image models and the visual cortex encode features in superposition. That is, we find non-axis aligned directions in the neural state space that are more interpretable than individual neurons. In addition, across both biological and artificial systems, we uncover the intriguing phenomenon of what we call \textit{feature synergy}\textemdash sparse combinations in activation space that yield more interpretable features than the constituent parts. Our work pushes in the direction of automated interpretability research for CNNs, in line with recent efforts for language models \citep{bills2023language, cunningham2023sparse, gurnee2023finding, bricken2022monosemanticity}. Simultaneously, it provides a new framework for analyzing neural coding properties in biological systems. Our results on neuroscience data add nuance to the concepts of \textit{disentanglement}, \textit{mixed selectivity}, \textit{representational drift} and \textit{representational universality} in the brain and suggest that insights gleaned from studying the coding properties of artificial neural networks may transfer to biological systems. These findings highlight potential synergy between mechanistic interpretability research and computational neuroscience, which may together reveal universal coding principles of neural representations.

%% file: sections/2-methods.tex
\section{Methods}\label{sec:methods}

We propose an approach for quantifying the interpretability of neural network activations that is grounded in human judgement, yet is fully automated and scalable. In general, individual neurons | i.e., $N$ directions corresponding to basis vectors of an activation space $\mathbb{R}^N$ | might not be interpretable. Yet, other directions in $\mathbb{R}^N$ might be: we refer to them as \textit{features}. For example, in Fig.~\ref{fig:project_overview} B) the human observer can define three directions that are interpretable and correspond approximately to horse-, car- and cat-like images. The superposition hypothesis stipulates that the activation space $\mathbb{R}^N$ of a neural network possesses several interpretable directions that are non-orthogonal \citep{elhage2022toy}. Given a CNN, we aim to identify such directions and quantify their interpretability through the following three steps: 

\begin{enumerate}
    \item \textbf{Collect neural network activations for a given dataset}. Images are passed through the network up to the layer under analysis, for convolutional layers, we average activations across space \citep{zimmermann2023scale} to generate a dataset in activation space $\mathbb{R}^N$.
    \item \textbf{Identify directions in activation space.} Directions may be provided by the neurons themselves (basis vectors) or by an algorithm (e.g., PCA, sparse coding, K-Means).
    \item  \textbf{Quantify the interpretability of each direction}. We compute an interpretability index (II) as the average pairwise similarity of the top $M$ Maximally Exciting Images (MEIs, defined in the next subsection) for each direction. Through a suite of experiments, we argue that the II is a meaningful measure of interpretability.
\end{enumerate}

\subsection{Quantifying Interpretability in Neural networks}

A neural network layer defines an activation space $f: \mathcal{X} \rightarrow \mathbb{R}^N$ with $N$ the number of neurons of that layer. We consider directions in this space, for example, individual neurons are represented as directions: the basis vectors of $\mathbb{R}^N$, i.e., for neuron $i$, $f_i(x)=f(x)e_i$ with $e_i$ the $i^{th}$ canonical basis vector and similarly $f_u(x)=f(x)u$ for any direction $u \in \mathbb{R}^N$. In activation space, some directions may be \textit{interpretable}, in the sense that they detect a single feature or concept within the image data. For example, an interpretable direction may detect features such as edges, corners, textures in early layers, or more abstract patterns in later layers such as dogs, cats, trucks. By contrast, other directions respond to several unrelated features or concepts. For instance, Fig.~\ref{fig:project_overview} (A) shows the first neuron firing in response to unrelated car- or cat-like images.

Maximally Exciting Images (MEIs) are defined as synthetic images that maximally activate a given direction in activation space ~\citep{erhan2009visualizing}. Given a direction $u$, we propose an \textbf{Interpretability Index (II)} computed as the average pairwise similarity of its top $M=5$ MEIs from a dataset of $D$ images, i.e., $f_u(x_1) \geq ... \geq f_u(x_M) \geq ... \geq f_u(x_D)$:
\begin{equation}\label{eq:interpretability-index}
    \mathrm{II}(u) = \frac{1}{M} \sum_{j=1}^M \sum_{k=1}^M \mathrm{sim}\Big(x_j, x_k\Big).
\end{equation}
In this work, we consider and compare several similarity metrics $\mathrm{sim}$, detailed below.

\subsection{Image Similarity Metrics}

We consider image similarity metrics that capture notions of similarity at different levels of abstraction: \textbf{1) Low-Level: Color} The color similarity between two images is defined as the difference between the average color (averaged across space, independently, for each color channel) in each image; \textbf{2) Mid-Level: LPIPS} Learned Perceptual Image Patch Similarity (LPIPS) \citep{zhang2018unreasonable} is a perceptual metric used for assessing the perceptual differences between images. It relies on a CNN such as VGG or AlexNet that has been pre-trained on an image classification task. Given two images, LPIPS extracts their respective feature maps from multiple layers of the CNN. LPIPS then computes the distance between the corresponding feature maps. The distances are scaled by learned weights and then aggregated to yield a single scalar value representing the perceptual similarity between the two images; \textbf{3) High-Level: Labels} The label similarity between two images is a value equal to 0 if the two images have been assigned different labels during a reference classification task, and equal to 1 if the two images have been assigned the same label. In our experiments, we use the CIFAR-10 dataset and associated classification task.

\subsection{From Human Psychophysics to In-Silico Psychophysics}\label{sec:insilico-psychophysics}

How can we validate whether the proposed interpretability index from Eq.~\ref{eq:interpretability-index} is indeed a sensible measure of interpretability? The concept of \textit{interpretability} is intimately tied to human judgment. A long history of theoretical inquiry has demonstrated the impossibility of identifying necessary and sufficient conditions for many natural semantic categories \citep{stekeler2012grundprobleme}.
Due to this difficulty, we adopt a pragmatic view, converting the question of whether a representation is interpretable into an empirical measure of the human interpretability judgment \citep{wittgenstein1953philosophische}.

\subsubsection*{Human Psychophysics} 

Psychophysics is an experimental paradigm for quantifying the relationship between stimuli (e.g. images) and the perceptions they produce for human observers. \cite{borowski2020exemplary} and \cite{zimmermann2023scale} have demonstrated that large-scale psychophysics experiments can be leveraged for conducting quantitative interpretability research. In these works, researchers used the judgments of human participants to quantify the interpretability of neurons in trained artificial neural networks. In \cite{zimmermann2023scale}, participants are shown 9 minimally and 9 MEIs for a given neuron. The participant is then asked to select one of two query images $x_1, x_2$ that they believe also strongly activates that neuron (see App.~\ref{sec:task_explanation} for an illustration). The \textit{(human) psychophysics accuracy} obtained for that neuron is defined as the percentage of participants that are able to select the correct image.

\subsubsection*{In-Silico Psychophysics} 

Psychophysics experiments provide a way of crowd-sourcing and quantifying human intuition of interpretability at scale. However, such experimentsts are time consuming, noisy and costly (\$12,000 for \citep{zimmermann2023scale}). Here, we propose a method for automating psychophysics experiments, with a model that faithfully approximates human judgments while requiring no human input. We replicate, \textit{in-silico}, the experiments of \cite{zimmermann2023scale}, comparing different similarity metrics as proxies for human judgments. In our experiments, the model computes the maximum similarity, according to the image similarity metric, between each of the query images $x_1, x_2$ and the set of MEIs. The model then chooses as its response the image that is the closest to that set

\begin{equation}
    \text{sim}(x, \text{MEI}(u))
    = \underset{k =1, ..., 9}{\operatorname{max}} \text{sim}(x, x_k),
\end{equation}
where $x_1, ..., x_9$ are the top $9$ MEIs for a neuron or direction $u$, and $\text{sim}$ the image similarity metric. The \textit{psychophysics accuracy} for a given neuron or direction $u$ is defined as the percentage at which the model selects the correct query image for that neuron, i.e.: 

\begin{equation}
        \text{Acc}(u) = \frac{\text{\# of correct selections for direction $u$}}{\text{\# of queries with direction $u$}}.
\end{equation}

We check in practice that directions $u$ with high interpretability index $\text{II}(u)$ are indeed more interpretable to a human observer. We note that we could have chosen the \textit{in-silico} psychophysics accuracy $\text{Acc}(u)$ of direction $u$ to quantify its interpretability: the more interpretable $u$ is, the easier it is for participants to correctly select images associated with it \citep{zimmermann2023scale}. However, we observe in practice that $\text{Acc}(u)$ is often at ceiling, and does not contain as much information as our proposed $\text{II}(u)$. Since it is also more expensive to compute, we use it only to validate the viability of the $\text{II}(u)$. Since the human psychophysics accuracy was computed only for directions corresponding to individual neurons, having in-silico psychophysics experiments is a key component of our approach.
Note that we chose to work with MEIs rather than feature visualisations \citep{simonyan2013deep, nguyen2019understanding}, because the latter has shown consistently lower interpretability in psychophysics studies \citep{borowski2020exemplary,zimmermann2021well,zimmermann2023scale}.

%% file: sections/3-results.tex
\section{Results}
\label{sec:results}

\subsection{Comparison: Human vs. In-Silico Psychophysics}
\label{sec:psychophysics}

\begin{figure}[H]
    \centering
    \includegraphics[width=0.99\textwidth]{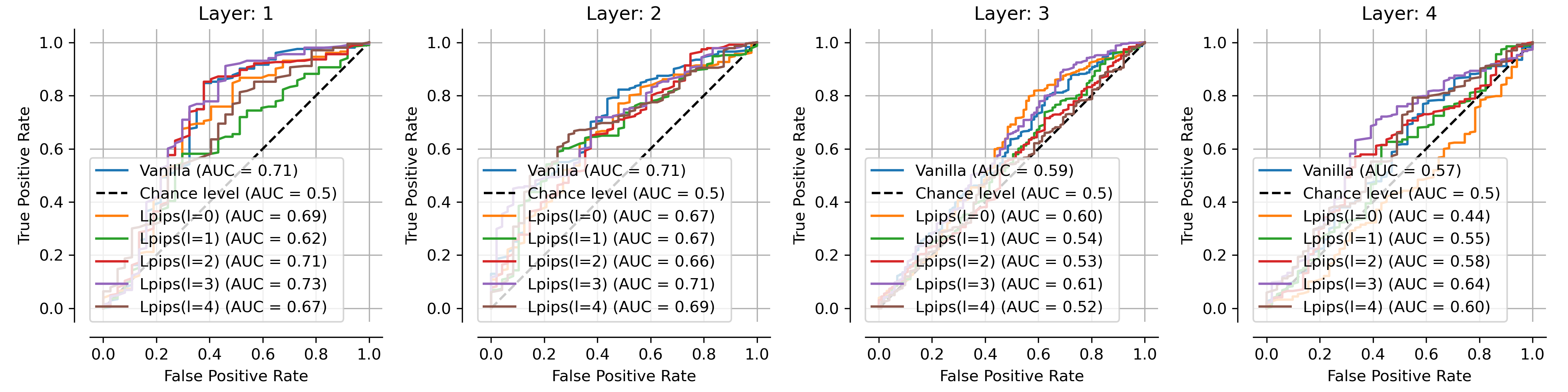}
    \caption{
    \textbf{Interpretability Metric vs. Human Behaviour.}
    Data from \cite{zimmermann2023scale}. Left to Right: Agreement between human and in-silico psychophysics on the predictability of the outputs of four `layers' (see text) within a ResNet50. Human and model agree on feature predictability for the ResNet50's early layers. For these layers, the proposed interpretability metric is a valid representation of the human's perception of interpretability. 
    AUC: Area Under the Curve. 
    }
    \label{fig:interpretability_metric}
\end{figure}

We first test the LPIPS similarity metric as a model of human perception of neuron interpretability. The experiments from \cite{zimmermann2023scale} quantify the interpretability of neurons in models trained on ImageNet-1k through crowd-sourced human perceptual judgments. We reproduce this experiment \textit{in-silico} by presenting the same image queries to a simple model based on the LPIPS metric. For each query, we evaluate whether the image selected by the model agrees with the image selected by human participants: it is a correct classification if they agree, incorrect otherwise. The results of this binary classification task are shown in Figure~\ref{fig:interpretability_metric} for the LPIPS image similarity metric \citep{zhang2018unreasonable}, on five of its AlexNet layers.

The predictions of the LPIPS model match human judgments well for earlier layers of the ResNet50\textemdash \textit{layers} 1 and 2\footnote{This refers to the PyTorch module names, corresponding to layers 10 and 23 in the network.}\textemdash with an AUC up to 0.71 (Figure~\ref{fig:interpretability_metric}, left two panels). While there is certainly room for improvement, we conclude that this metric, based on LPIPS-based pairwise image comparison, serves as a good first proxy of human perception of interpretability. Crucially, our metric has the added benefit of not having to recruit a cohort of human participants. Thus, we will use this metric to evaluate the interpretability of features across neural network layers in the next subsections. Since the interpretability metric is more accurate for early layers, we focus the remainder of our analyses on layer 1 of the same ResNet50 architecture trained on CIFAR-10.

\subsection{Identifying Interpretable Directions in Feature Space}

We next apply the II to analyze the interpretability of features in a ResNet50 pre-trained on the CIFAR-10 image dataset \citep{krizhevsky2009learning}\footnote{Hosted at \href{https://github.com/huyvnphan/PyTorch_CIFAR10}{https://github.com/huyvnphan/PyTorch\_CIFAR10}}. We evaluate several methods for identifying interpretable directions in activation space: PCA, ICA, NMF, $K$-Means with cosine similarity, and the shallow sparse autoencoder used in \cite{conjecture2022sparse_ae} (see Appendix \ref{sec:sparse_autoencoder} for sparse AE analyses). We evaluate several similarity metrics and compute the II for each, comparing the interpretability of individual neurons in a layer with the interpretability of identified directions from that layer.

\begin{figure}[H]
    \centering
    \includegraphics[width=0.99\textwidth]{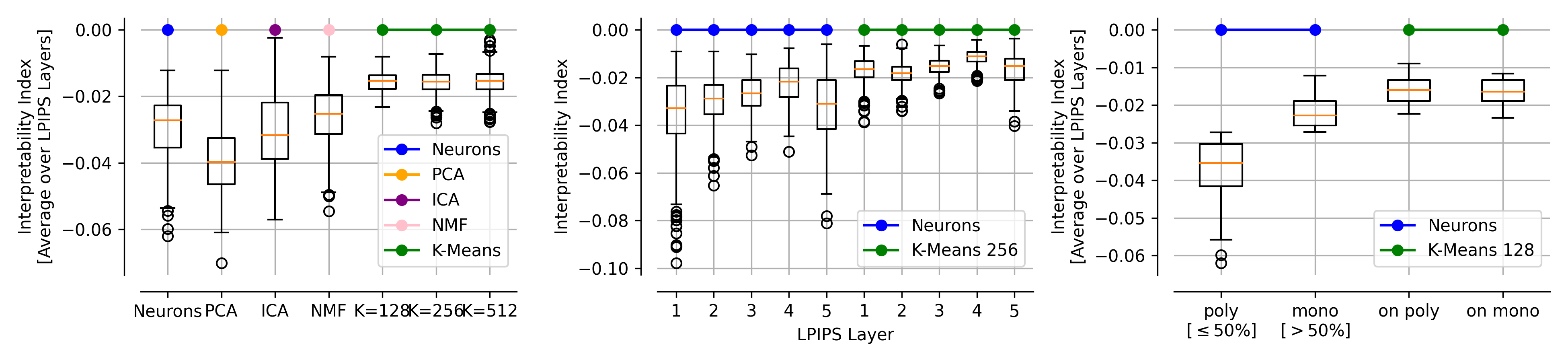}
    \caption{
    \textbf{Quantification of Interpretability.}
    Left: II score [a.u.] distribution for neurons ($N=256$), PCA, ICA and NMF baselines, and K-Means as a function of $K \in \{128, 256, 512\}$.
    Middle: II score distribution for neurons ($N=256$) and K-Means ($K=256$) as a function of LPIPS layer.
    Right: II score distribution for uninterpretable neurons ($N=128$), i.e. those with II score below the median, and interpretable neurons ($N=128$), i.e., those with II score above the median; II score distribution for K-Means ($K=128$) computed on each of these subsets separately.
    }
    \label{fig:fig_mis}
\end{figure}

For a quantitative comparison, we present comparative box plots for the distributions of II indices for neurons and directions in Figure~\ref{fig:fig_mis} while we vary: the LPIPS layer defining the similarity metric used in the II (left), the number of $K$ directions in activation space (middle), and distributions after splitting neurons into uninterpretable and interpretable groups (right). We observe that the $K$-means approach detects directions that are indeed more interpretable (higher II) than the individual neurons of the activation space | independently of the LPIPS layer considered for the II (Figure~\ref{fig:fig_mis} left). Our method achieves higher II values ($\text{mean}=-0.0159$) than all baselines and the sparse autoencoder (best $\text{mean}=-0.0188$) (detailed comparison in App.~\ref{sec:sparse_autoencoder}) [note that the II has arbitrary units]. Interestingly, the number of directions $K$ does not impact their IIs in the regime tested (Figure~\ref{fig:fig_mis} left). Thus, we focus our analyses on the $K=N=256$ setting for a fair comparison.

The $K$-means approach can detect directions within subsets of uninterpretable neurons as well as within interpretable neurons, as we do not observe II differences in Figure~\ref{fig:fig_mis} (right). Further, transforming interpretable neurons and uninterpretable neurons into their direction increases the II of both. We see a trend where the II increases with respect to the LPIPS layer used, which is a similar pattern as we saw in Figure~\ref{fig:interpretability_metric}. 

\begin{figure}[H]
    \centering
    \includegraphics[width=.99\textwidth]{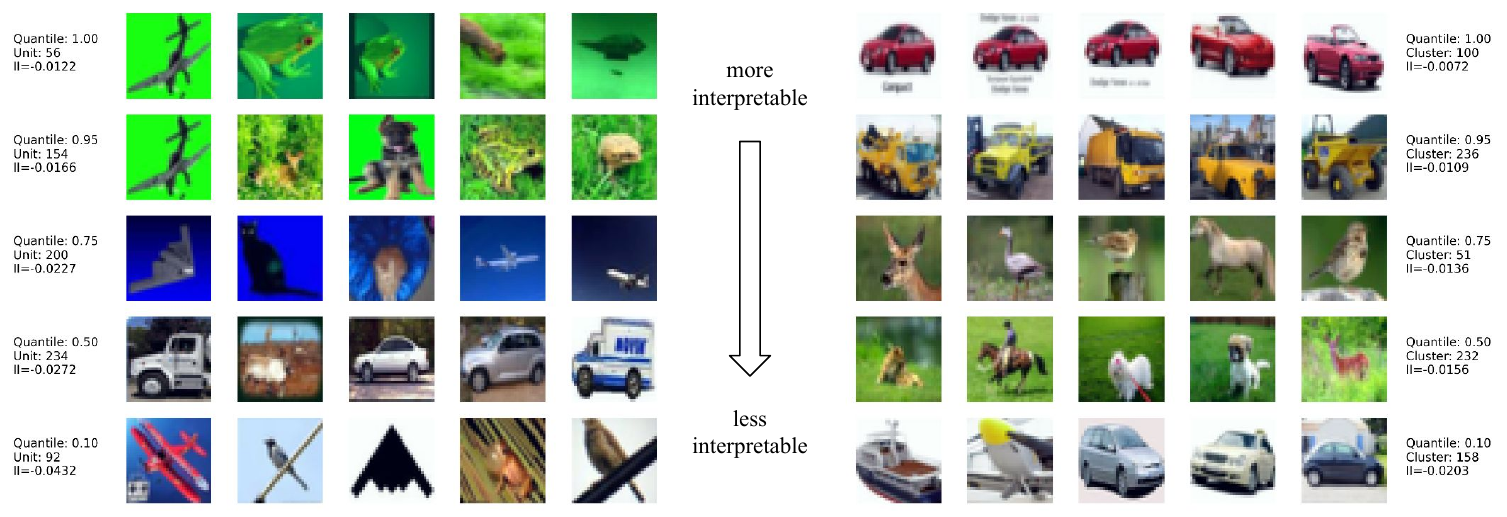}
    \caption{
    \textbf{MEIs of Neurons and Interpretable Directions.}
    These are the Maximally Exciting Images (MEIs) for neurons (left) and directions (right) as retrieved with $K$-Means.
    To represent the interpretability index (II) distribution, we show neurons and directions at different II quantiles. \vspace{-0.4cm}
    }
    \label{fig:fig_meis}
\end{figure}

For a qualitative comparison, Figure~\ref{fig:fig_meis} shows the Maximally Exciting Images (MEIs) for 5 neurons (left) and 5 directions extracted from $K$-Means (right) selected in 5 different quantiles of II values (to avoid cherry picking in this qualitative comparison). The distributions of II indices is shifted towards the higher values for the directions detected by $K$-means, as shown by the II values associated with each quantile. This is confirmed by the visualization of MEIs, which appear more visually coherent to the human observer for the directions (right) compared to the neurons (left).

\subsection{Comparing Similarity Metrics for In-Silico Psychophysics}

\begin{figure}[H]
    \centering
    \includegraphics[width=0.99\textwidth]{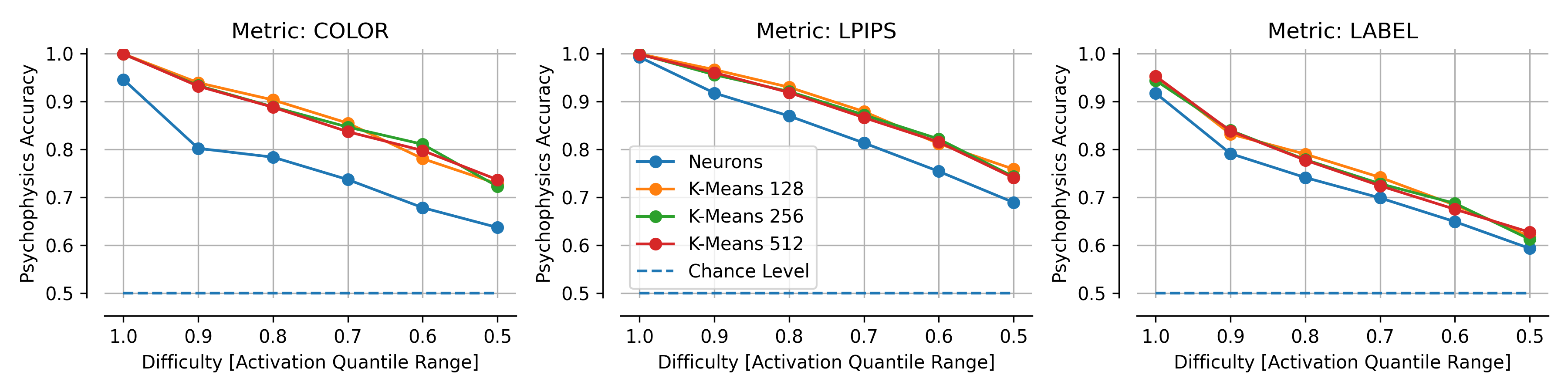}
    \caption{
    \textbf{\textit{In Silico} Psychophysics Performance.}
    Accuracy across neurons and interpretable directions revealed by $K$-Means clusters ($K \in \{128, 256, 512\}$) for \textit{in silico} psychophysics task for different levels of difficult, i.e., limiting query and reference image selection to the central range of activations (e.g., from the $0.45^{th}$ until the $0.55^{th}$ quantile, see \citep{zimmermann2023scale}).
    Predictions are made based on different metrics from low level semantics (colour match, left), over mid level semantics (LPIPS average over layers, center), to high level semantics (label match, right).
    }
    \label{fig:fig_psycho}
\end{figure}

We now compare the interpretability of the directions measured using the three image similarity metrics described in Section~\ref{sec:methods}. Each metric defines similarity at a different level of abstraction, from low-level to high-level: same \textit{color} (Figure~\ref{fig:fig_psycho} left), same \textit{perceptual structure} as defined by LPIPS (Figure~\ref{fig:fig_psycho} middle) or same \textit{category} (Figure~\ref{fig:fig_psycho} right). For each metric, we perform the \textit{in-silico} psychophysics task from Section \ref{sec:psychophysics}, varying the difficulty of the psychophysics experiment. The difficulty of a task is controlled by choosing query images that cause less extreme activations\textemdash i.e. are farther away from the set of MEIs \citep{borowski2020exemplary}. This allows us to probe a more general understanding of the interpretability of a neuron or direction instead of limiting our analyses to the most preferred stimuli \citep{vinken2023neural}.

As expected, we see in Figure~\ref{fig:fig_psycho} that both the neurons and the directions have a decreased psychophysics accuracy as the task becomes more difficult. The directions detected by our approach are more predictable than the individual neurons across low, mid and high-level semantics and across task difficulties. The largest improvement over individual neurons is observed for the low-level semantics using colors, and the improvement decreases as we move towards higher level semantics. Additionally, as observed in Figure~\ref{fig:fig_mis}, the number of clusters $K$ does not impact the accuracy.

Lastly, in recently published work, \cite{bricken2022monosemanticity} test whether observed interpretability in activation space is a function of the model or the data\textemdash that is, they test whether untrained models possess non-axis aligned directions that are more interpretable than individual neurons and find evidence that they do. 
We perform the same experiment, running our analysis on untrained versions of the models we analyze here, and find that there is indeed\textemdash even before training\textemdash a gap in interpretability between neuron axes and activation clusters (see Appendix G).
This aligns with prior work on the expressive power of untrained CNNs \citep{frankle2020training} and suggests paths for further investigation.

\subsection{Pairwise Synergies Between Neurons}

\begin{figure}[H]
    \centering
    \includegraphics[width=0.99\textwidth]{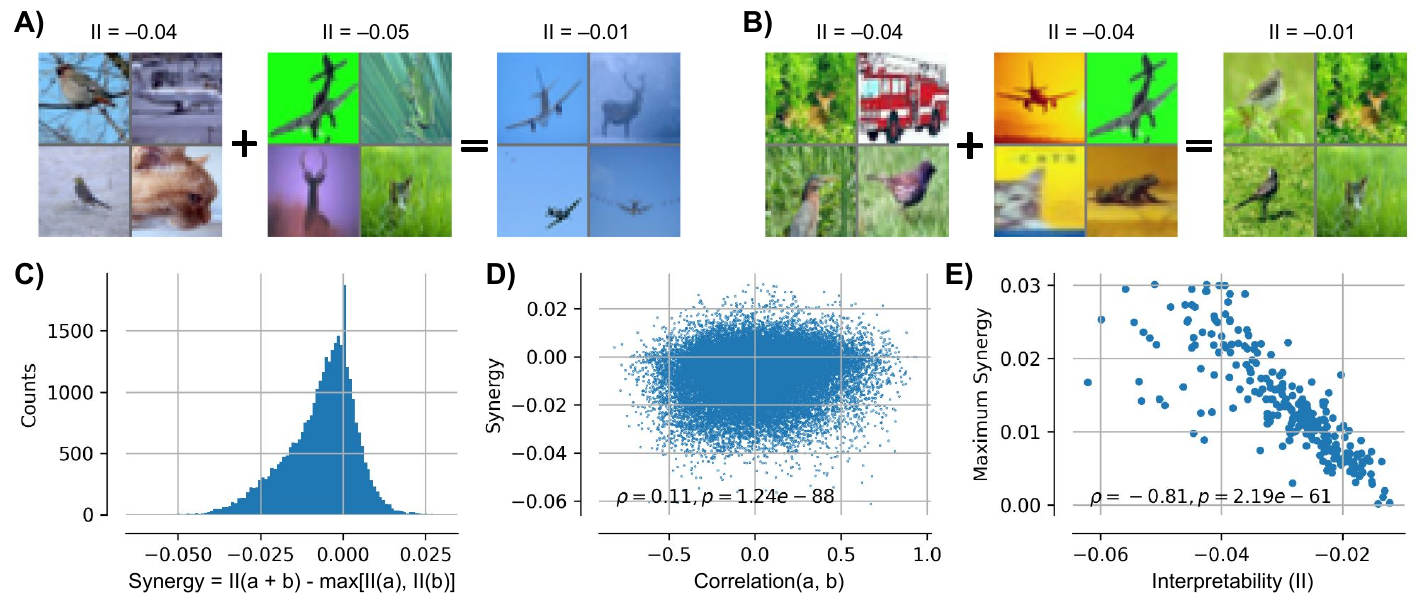}
    \caption{
    \textbf{Synergies.}
    \textbf{A), B)} Example pairs (two highest synergies) of neurons and the result when adding them (all visualized by their 4 MEIs).
    \textbf{C)} Histogram of synergies for every pair of neurons.
    \textbf{D)} A slight positive relationship between the correlation and synergy over all pairs of neurons (i.e., more correlated neuron pairs have higher synergies).
    \textbf{E)} A strong negative relationship between the II of a neuron and the maximum synergy it can achieve (i.e., pairings dilute interpretable neurons).
    }
    \label{fig:fig_synergy}
\end{figure}

Efficient coding principles such as minimal wiring length \citep{laughlin2003communication}, as well as the circuit analysis approach of mechanistic interpretability \citep{conmy2023towards,nanda2023progress} inspire us to look for minimal subcircuits that increase interpretability.
Specifically, we investigate the synergies between pairs of neurons. For all pairs of neurons $a, b$ in the same ResNet50 layer, we compute the II score for their added (z-scored) activity. The \textit{synergy} is the difference between this II score and the maximum of their individual II scores to account for pairings with highly interpretable neurons:
\begin{equation}
    \text{Synergy}(a, b) = \text{II}(a+b) - \text{max}\left[ \text{II}(a), \text{II}(b)\right].
\end{equation}
The synergy measures whether adding these neurons produces a direction in activation space that is more interpretable than taking each neuron individually. This is visualized in Figure~\ref{fig:fig_synergy} A) and B) which show two pairs of neurons $a, b$ with the highest synergy: the MEIs resulting from their addition are more interpretable  that their individual MEIs.

The histogram of Figure~\ref{fig:fig_synergy} C) shows a large fraction of negative values of the synergy, i.e., most pairings are, as expected, detrimental for interpretability.
However, a good fraction of the added neurons $a+b$ become more interpretable. Figure~\ref{fig:fig_synergy} D) shows that correlated neurons tend to have higher synergy but correlation alone does not explain everything: two neurons can be uncorrelated, yet their addition can produce a very interpretable feature. This shows that our notion of interpretability is distinct from the familiar notion of decorrelation. Lastly, we find that more interpretable neurons (higher II) show lower maximal synergy (Figure~\ref{fig:fig_synergy} E)). This suggest that their representation is already interpretable and that any pairing would only dilute it.

\subsection{Application to Biological Neural Data}
Findings of \textit{mixed selectivity}, i.e., hard to interpret neurons that code for multiple unrelated features have been reported before in neuroscience \citep{yoshida2007odorant,rigotti2013importance,fusi2016neurons}.
This suggests that the cortex may also encode meaningful features in superposition. 
Below, we perform the same analysis as above, but for cortical recordings from inferior temporal (IT) visual cortex in macaque monkeys\textemdash a cortical area involved in high level visual object recognition \citep{hung2005fast} with a specific preference for faces \citep{tsao2006cortical}.

\subsubsection{Face Cell Responses}

\begin{figure}[H]
    \centering
    \includegraphics[width=0.99\textwidth]{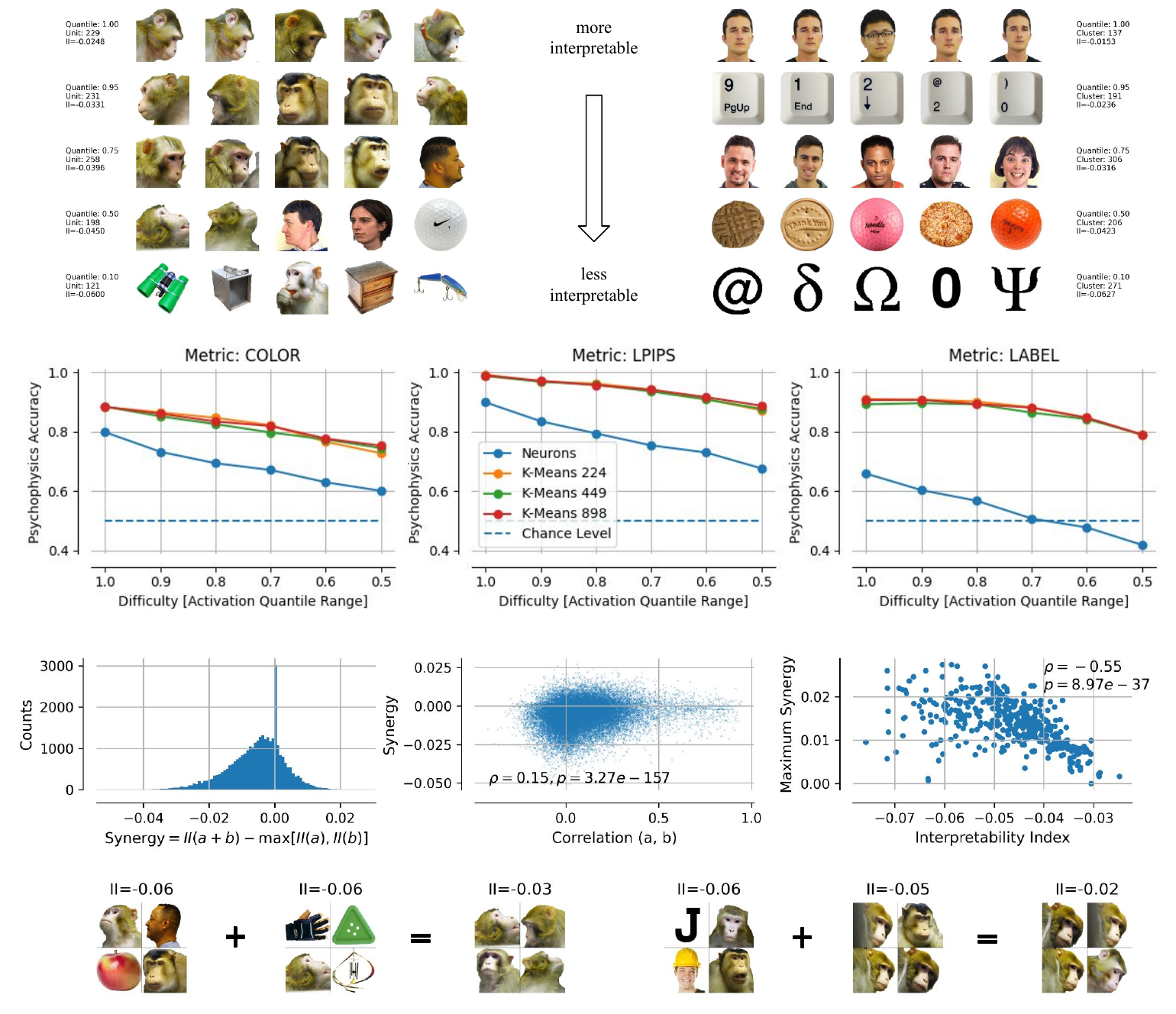}
    \caption{
    \textbf{Face Cell Responses in IT Cortex.}
    \textbf{\nth{1} row)} Maximally Exciting Images (MEIs) for neurons (left) and directions (right) as retrieved with $K$-Means (see Fig.~\ref{fig:fig_meis}).
    \textbf{\nth{2} row)} Accuracy across neurons and interpretable directions revealed by $K$-Means clusters for the \textit{in silico} psychophysics task for different levels of difficulty (see Fig.~\ref{fig:fig_psycho}).    
    \textbf{\nth{3} row)} (left to right, see Fig.~\ref{fig:fig_synergy}) Histogram of synergies for each pair of neurons; Relationship between the correlation and synergy over all pairs of neurons; Relationship between the II of a neuron and the maximum synergy it can achieve.
    \textbf{\nth{4} row)} Example pairs (two highest synergies) of neurons and the result when adding them (see Fig.~\ref{fig:fig_synergy}). 
    }
    \label{fig:face_cells}
\end{figure}

We first examine a dataset from \citet{vinken2023neural}, consisting of the responses of $449$ neurons (sites) to 1379 images (447 faces 932 non-face objects).
We perform the same analysis pipeline as for the CNN above, i.e., we clustered the activations using K-Means and studied the learned \textit{features} by their MEIs and our \textit{in silico} psychophysics.
The results are shown in Fig.~\ref{fig:face_cells}.
The top row is the same as Fig.~\ref{fig:fig_meis}, showing quantiles of II sorted neurons/features.
This is a rather striking demonstration of the previous claim that IT cortex represents a `domain-general object space' \citep{vinken2023neural}, i.e., we find highly interpretable activation clusters that code, e.g., for faces, keyboard keys, round objects or characters.
In the center row of Fig.~\ref{fig:face_cells}, we perform the same \textit{in silico} psychophysics experiment as above (labels are now provided by $3$ distinct image conditions, see \citep{van2008visualizing}).
Intriguingly, we find the same if not a larger effect of increased interpretability when moving from individual neurons to the features we find with K-Means.
Lastly, in the bottom row of Fig.~\ref{fig:face_cells}, we perform the same synergy experiment as in Fig.~\ref{fig:fig_synergy} and also find the same qualitative pattern, including a skewed distribution over synergies, a positive link with pairwise neural correlations, and a negative link with the II score.

These results are an interesting extension of the conclusions from the original \cite{vinken2023neural} paper, in which the authors concluded that MEIs give an incomplete picture and that face cells should rather be understood as representing a domain-general object space. 
We fully agree with the former conclusion\textemdash when limited to individual neurons. 
The additional insight that we obtain here is that the object space, represented by multiple IT neurons, is spanned by groups of features (in \textit{superposition}) whose MEIs are meaningful in the sense that they correspond to interpretable coding directions (Fig.~\ref{fig:face_cells} (\nth{1} row)) and whose activations are interpretable across a wide range of quantiles (Fig.~\ref{fig:face_cells} (\nth{2} row)).

\subsubsection{Disentangling Interpretable Features in IT Cortex}

\begin{figure}[H]
    \centering
    \vspace{0.8cm}
    \includegraphics[width=0.99\textwidth]{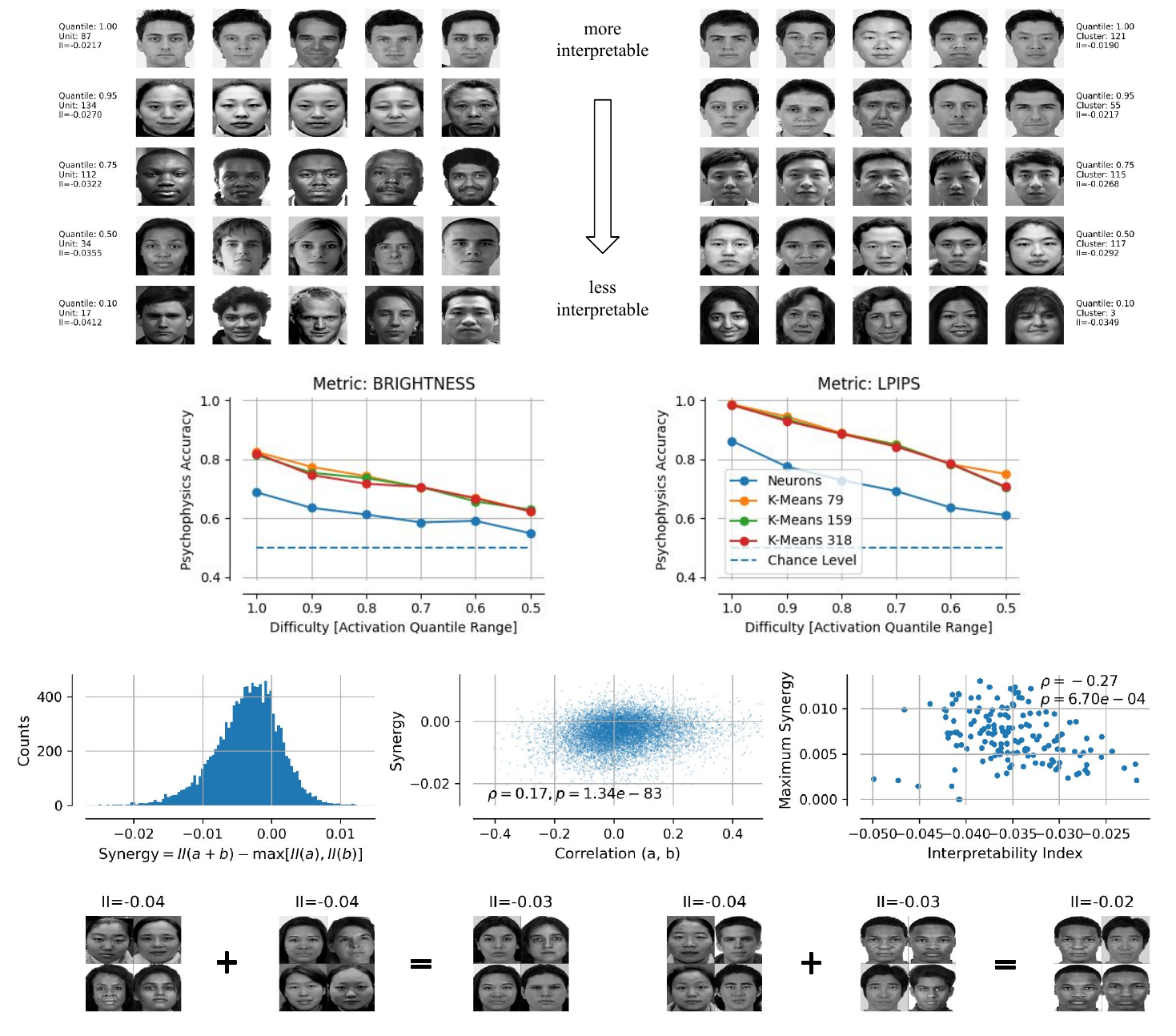}
    \caption{ 
    \textbf{Interpretability in IT Cortex.}
    \textbf{\nth{1} row)} MEIs for neurons (left) and directions (right) as retrieved with $K$-Means (see Fig.~\ref{fig:fig_meis}).
    \textbf{\nth{2} row)} Accuracy across neurons and interpretable directions revealed by $K$-Means clusters for \textit{in silico} psychophysics task for different levels of difficult (see Fig.~\ref{fig:fig_psycho}).    
    \textbf{\nth{3} row)} (left to right, see Fig.~\ref{fig:fig_synergy}) Histogram of synergies for every pair of neurons; Relationship between the correlation and synergy over all pairs of neurons; Relationship between the II of a neuron and its maximal synergy.
    \textbf{\nth{4} row)} Example pairs (highest synergies) of neurons and the result when adding them (see Fig.~\ref{fig:fig_synergy}). \vspace{0.4cm}
    }
    \label{fig:higgins}
\end{figure}

Next, we apply this analysis to the dataset from \citet{higgins2021unsupervised}, which consists of 159 neurons in anterior middle (AM) macaque face area that were presented with 2100 human and monkey face images. We apply the same analysis pipeline as before, i.e., we cluster the activations using K-Means and study the learned \textit{features} through their MEIs and \textit{in silico} psychophysics. The results are shown in Fig.~\ref{fig:higgins}.Note that the images used in the experiment were greyscaled. Thus, we are limited to considering only brightness rather than color for the low-level metric. For the mid-level LPIPS metric, we feed the greyscale value into all three colour channels. For this dataset, it is not possible to consider the label-based method, as there is no category information provided for the images in this dataset.

Again, the MEIs Fig.~\ref{fig:higgins} (top) and \textit{in silico} psychophysics Fig.~\ref{fig:higgins} (center) tell a consistent story. That is, we find a significant increase in interpretability (psychophysics performance across all levels of difficulty) when moving from individual neurons to the K-Means features.
Lastly, the synergy experiment Fig.~\ref{fig:higgins} (bottom) also shows the same result pattern with a skewed synergy distribution, a positive link between pairwise neural correlations and synergies, and a negative link between interpretability and maximal synergy.

\begin{figure}[H]
    \centering
    \includegraphics[width=0.99\textwidth]{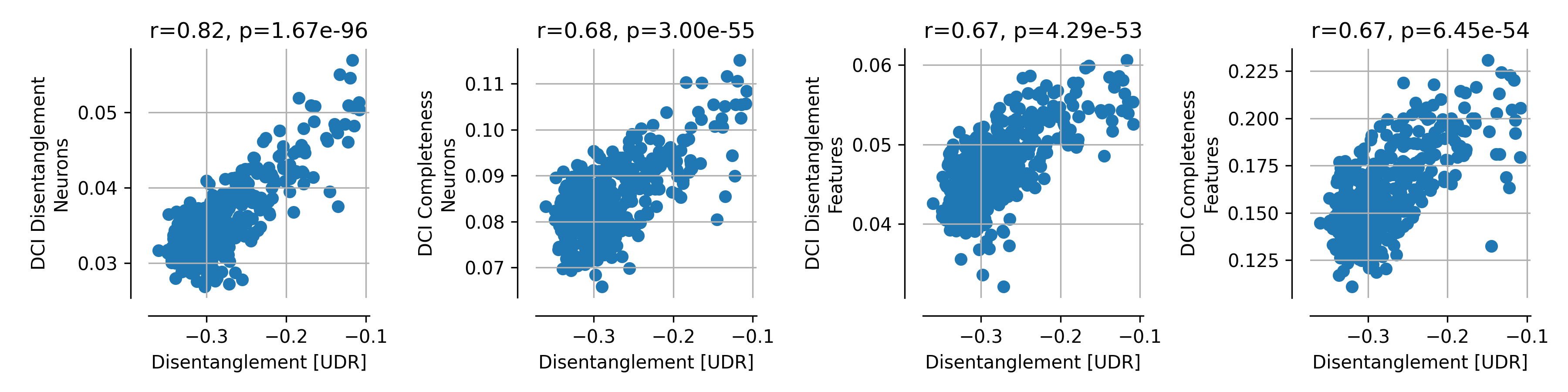}
    \includegraphics[width=0.9\textwidth]{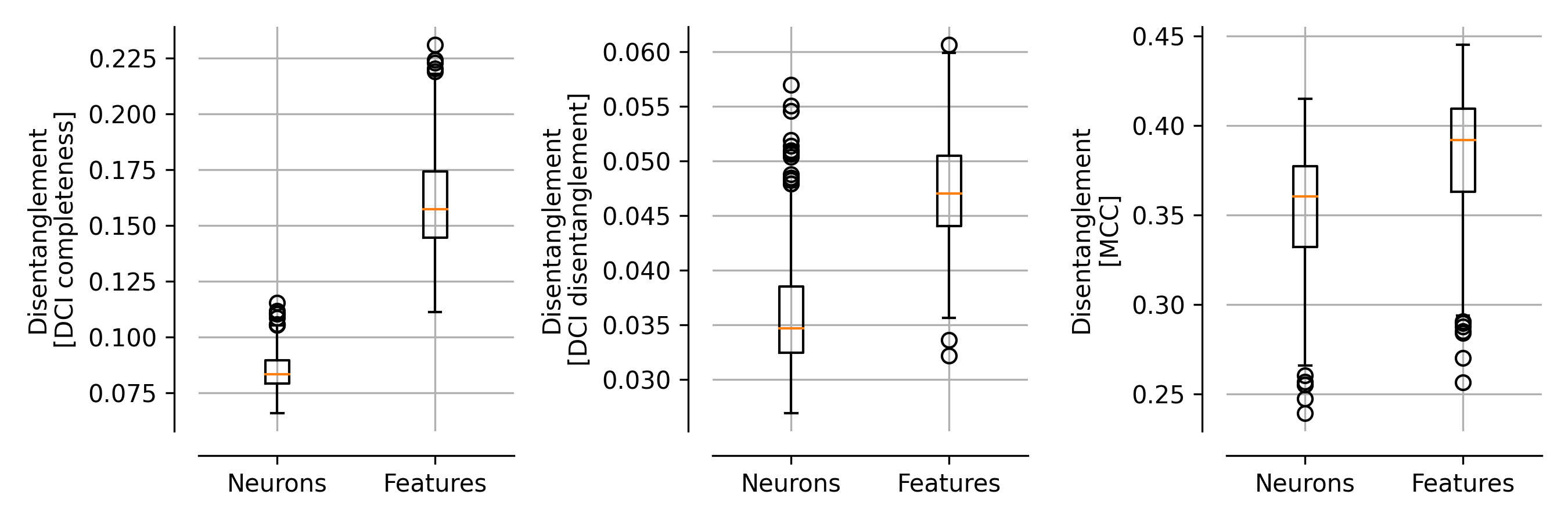}
    \includegraphics[width=0.8\textwidth]{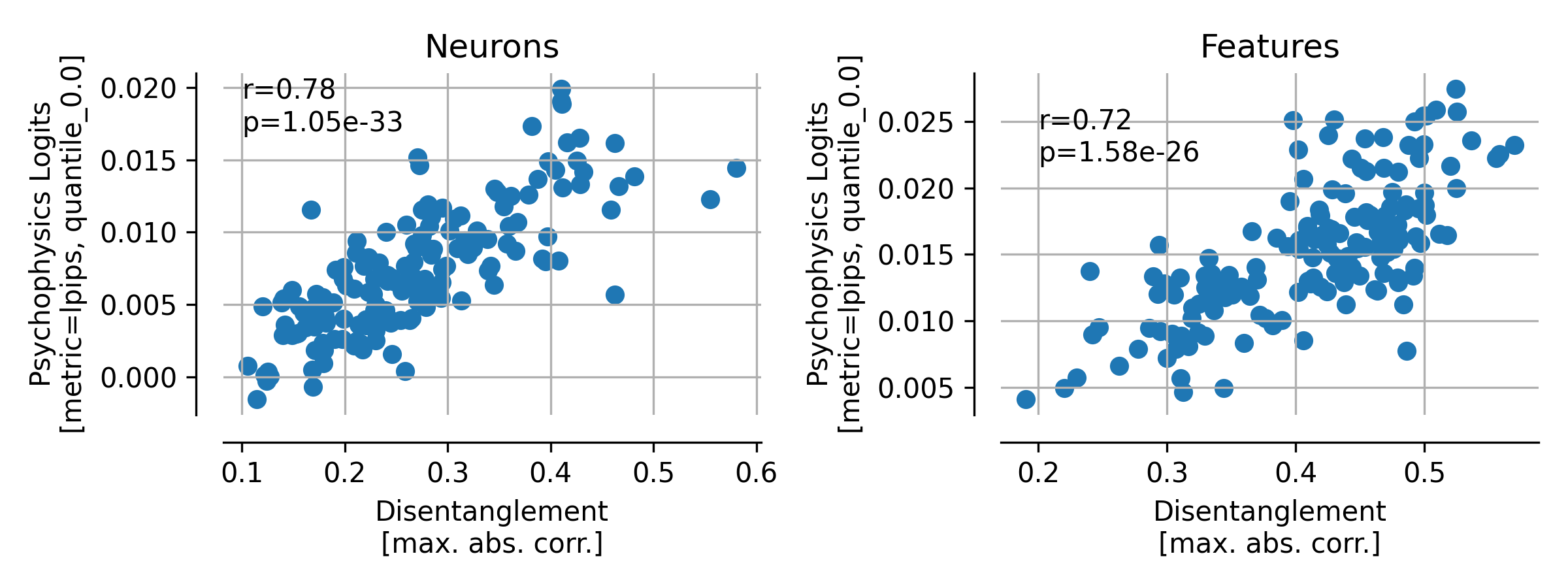}
    \caption{
    \textbf{Disentanglement and Interpretability in IT Cortex.}
    \textbf{Top)} Relation between \textit{unsupervised} disentanglement \citep{duan2019unsupervised} of 400 models ($\beta$-VAE) and \textit{supervised} (models vs.\ neurons / features) disentanglement of neurons and features \citep{higgins2021unsupervised}.
    \textbf{Center)} Different measures of disentanglement (\textit{DCI} \citep{eastwood2018framework} and \textit{MCC} \citep{hyvarinen2016unsupervised,hyvarinen2017nonlinear}) for neurons and features.
    \textit{DCI Disentanglement} corresponds to the metric named \textit{alignment} in the original paper \citep{eastwood2018framework,higgins2021unsupervised}.
    \textbf{Bottom)} Relationship between interpretability (psychophysics logits for LPIPS metric and full quantile range, i.e., $x=1.0$ in Fig.~\ref{fig:higgins}, \nth{2} row, left) and disentanglement (see text). 
    }
    \label{fig:higgins_dis_vs_psy}
\end{figure}

In the original paper by \cite{higgins2021unsupervised}, the key insight was that \textit{individual} neurons encode disentangled, interpretable features of the data.
By contrast, we find that directions in activation space that mix multiple neurons are more interpretable than individual units.
To gain more insight into this discrepancy, we perform a similar analysis of disentanglement as the original paper.
In the original paper, they trained $400$ $\beta$-variational autoencoder ($\beta$-VAE) models \citep{higgins2016beta} with different seeds and hyper-parameters that, empirically, find interpretable factorizations of the data \citep[although see][]{hyvarinen1999nonlinear,locatello2019challenging}.
They used an unsupervised metric of \textit{disentanglement} \citep{duan2019unsupervised} to check if more disentangled models have a better one-to-one correspondence with IT neurons, and found a positive relationship.\footnote{Note that the same desideratum of having a sparse readout that links model features with neural responses leads to the identification of functional cell types in neural system identification \citep{klindt2017neural,ustyuzhaninov2019rotation}.}
We find the same positive relationship in Fig.~\ref{fig:higgins_dis_vs_psy} (top) for both neurons (left) and K-Means features (right).

In the middle of Fig.~\ref{fig:higgins_dis_vs_psy}, we report the distribution over different disentanglement metrics for neurons and features.
Surprisingly, we find that the \textit{features} tend to achieve higher scores (across the $400$ model instances), suggesting that they are more disentangled than individual neurons.
This finding is particularly interesting because \textit{disentanglement} and \textit{interpretability} are logically separable concepts\textemdash the former can be mathematically formalized as \textit{source recovery} in (non-)linear ICA \citep{hyvarinen2016unsupervised,hyvarinen2017nonlinear,klindt2020towards,hyvarinen2023identifiability}, while the latter is a complex function of human semantics.
However, based on these results, we hypothesize that our measures of interpretablity are in fact related to classical notions of disentanglement or \textit{source recovery}. Further supporting this idea, in the bottom of Fig.~\ref{fig:higgins_dis_vs_psy}, we find a strong relationship between a supervised measure of disentanglement (i.e., the maximal absolute correlation between a neuron/feature and the model units, for the model with the highest UDR score) and our interpretability score (in terms of psychophysics logits).\footnote{We could not use the same disentanglement metrics above since those are across models, while here, we report scores across neurons/features.} 

The results of these analyses are interesting for two reasons. First, they provide an alternative interpretation of the original data: Neurons may sometimes align with disentangled factors of the data, however, activity clusters that involve multiple neurons tend to be even better aligned with disentangled models. Second, we find that our measures of interpretability (here \textit{in silico} psychophysics accuracy) is strongly related to the more mathematically-grounded concept of disentanglement \citep{hyvarinen2023identifiability}.

\subsubsection{Universality and Representational Drift}

Shifting the focus away from individual neurons and, instead, considering meaningful clusters of activity in state space the computational primitives of neural network function provides a fresh view on \textit{universality} \citep{olah2020overview}, i.e., the similarity of representations of different neural systems, and \textit{representational drift}, i.e., the changes of representations within the same neural system over time \citep{driscoll2022representational}.
To test this hypothesis, in this section we examine a neuroscience dataset from \citet{allen2022massive}. This dataset is interesting because it allows us to study representations across brain areas, across time and across human subjects.
Specifically, this dataset consists of functional magnetic resonance imaging (fMRI) recordings of $8$ human subjects, in over $16$ brain areas, viewing $9,000-10,000$ natural images across $30-40$ scan sessions \citep[for more details, see]{allen2022massive}.
The recorded \textit{units} (i.e., voxels/cubes in a $3D$ sampling grid over the human brain) combine the activity of many neurons across time which presents an interesting extension to the hypothesis of superposition, since these voxels can already be considered linear combinations of individual neurons with, potentially, interesting filtering properties \citep{kriegeskorte2010does}.

\begin{figure}[t]
    \centering
    \includegraphics[width=0.99\textwidth]{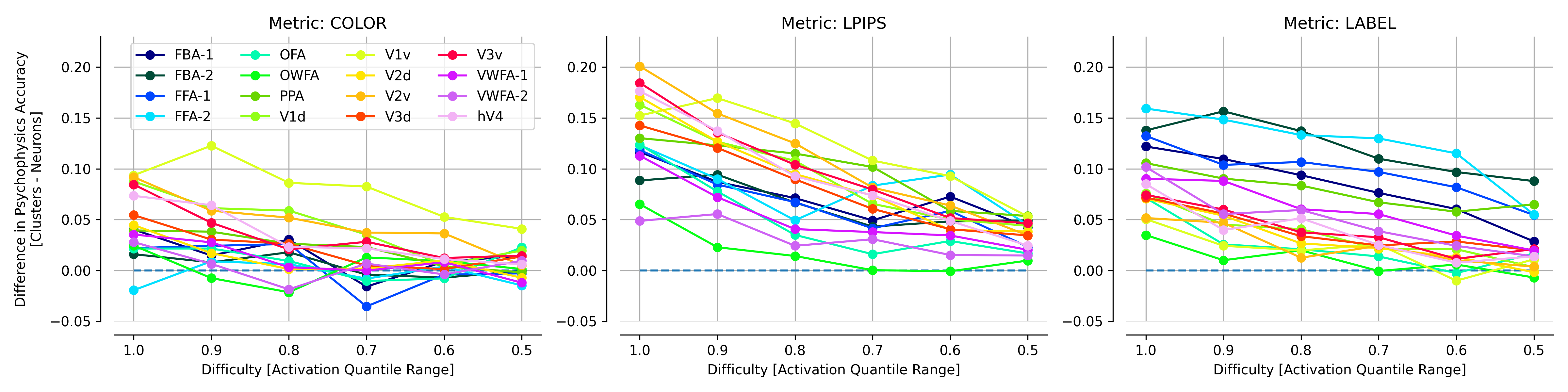}
    \includegraphics[width=0.99\textwidth]{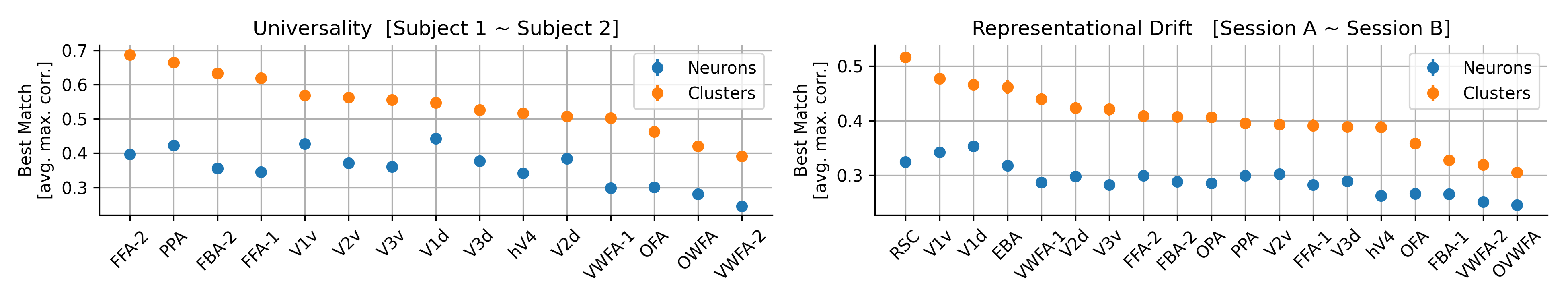}
    \includegraphics[width=0.99\textwidth]{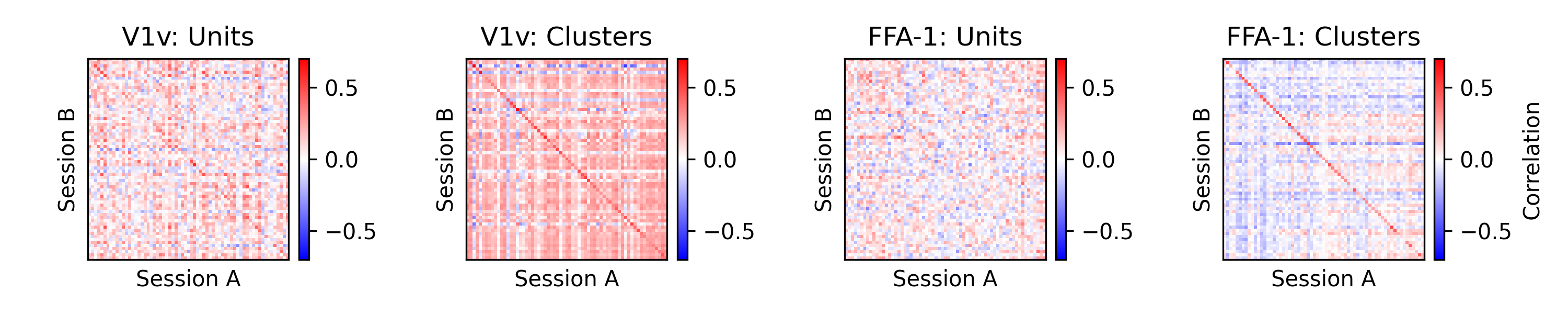}
    \caption{
    \textbf{Interpretable Features Transfer Across Subjects and Recording Sessions.}
    Using the dataset from \cite{allen2022massive}, we examine: \textbf{Top)} Differences in psychophysics accuracy between clusters and units (fMRI voxels) for different difficulties (horizontal axis), metrics (columns) and brain areas (colours). A value above zero indicates that the clusters achieve a higher psychophysics accuracy, i.e., are more interpretable.
    \textbf{Middle, left)} Best matching units (maximal correlation between activations, for $N=1000$ images) across all brain areas of distinct subjects\textemdash testing representational universality across brains \citep{vinken2023neural}.
    \textbf{Middle, right)} Same as middle left but across scan sessions\textemdash testing representational drift across time (for $N=88$ images that were presented in both scan sessions $13$ and $14$).
    \textbf{Bottom)} Exemplary low (left, V1v) and high (right, FFA-1) level representation drift, i.e. correlation of units and clusters across sessions.
    }
    \label{fig:zvi}
\end{figure}

As in the previous experiments, we perform the same analysis pipeline of identifying clusters in neural activity space and calculating the psychophysics accuracy for clusters and units. We find, again, that for all brain areas (Subject 1) the clusters achieve a higher psychophysics accuracy, suggesting that they are more interpretable than the individual units Fig.~\ref{fig:zvi} (top). Moreover, we see a tendency for lower areas (e.g., V1v, \citep[see ][]{allen2022massive}) to have a larger gain for the low level color metric, whereas higher areas (e.g., FBA-2, FFA-2) have a larger gain for the high level label metric. This provides further support for the existence of non-axis aligned interpretable features across visual cortical areas.

A key finding in interpretability research is the phenomenon of \textit{convergent learning} \citep{li2015convergent}\textemdash the observation that diverse neural network architectures trained on similar tasks converge to similar internal representations. This naturally leads to the question of \textit{universality} \citep{olah2020overview}: Is there a set of canonical features for a data domain that consistently emerge across neural systems? Here, we ask whether the interpretable directions identified with out method exhibit greater universality than neurons alone. If this hypothesis is true, we would expect to see greater similarity between \textit{directions} in neural activity space across subjects than between \textit{neurons}.
We next test the ``universality'' of these features, by examining how well the discovered directions transfer across subjects and recording sessions.\footnote{Note that clusters can be easily transferred across recordings or subjects by computing the centroids in the target space based on the clustering labels in the source space.}

In Fig.~\ref{fig:zvi} (middle), we find that the identified directions transfer across subjects (middle, left) better than units alone. In addition, we find that these directions transfer better across recording sessions (middle, right) than individual units. In Fig.~\ref{fig:zvi} (bottom), we show exemplary low (left) and high (right) level brain areas where the activity across sessions is much more correlated for the clusters than for the units. This is in line with prior results, where \citet{roth2023representations} observed that the representational dissimilarity (i.e., the relative position of activity patterns for different stimuli \citep{kriegeskorte2008representational} remained stable across scan sessions. This suite of findings provides support to the idea that more interpretable features transfer more easily across different representations, in line with the findings of \citet{vinken2023neural} about the universality of interpretable features.

%% file: sections/4-discussion.tex
\section{Discussion}

In this work, we have proposed a quantitative metric of \textit{interpretability} and a method for finding interpretable features in activation space. We hope that further research will find better metrics and better feature identification methods.
Nevertheless, we believe that our initial combination of metric and feature recovery method used here demonstrates the viability of our framework for automating interpretability research for vision models and visual cortex. In particular, we emphasize the value of validating quantitative metrics of interpretability against large-scale human psychophysics experiments of interpretability \citep{zimmermann2023scale}. This allows us to scale human intuition to large-scale, complex neural network models\textemdash thus automating what is ordinarily done in mechanistic interpretability research by hand \citep{leavitt2020towards}. We hope that this approach will ultimately lead to a better understanding of neural coding principles and cast light into the black box of deep network representations. 

Shifting focus from individual neurons to populations has been an important development in neuroscience \citep{averbeck2006neural,stanley2013reading,hebb2005organization,gao2015simplicity,jacobs2009ruling,ebitz2021population}.
In fact, \textit{mixed selectivity} is widely observed in neuroscience, \citep{yoshida2007odorant,rigotti2013importance} and there are coding advantages believed to be conferred by such a representation \citep{fusi2016neurons,driscoll2022representational}.
We also tested a recent neural coding hypothesis that combines sparse coding with disentanglement in the framework of the \textit{sparse manifold transform} \citep{chen2018sparse}.
In App.~\ref{sec:smt} we find support for the notion that interpretable features are more sparsely localized on the data manifold.
Moreover, this theoretical framework could help further elucidate the link between interpretability (of discrete clusters) and disentanglement that we found in neural data \citep{higgins2021unsupervised}.
Lastly, such a code may be more robust to input perturbations \citep{morcos2018importance}, as suggested by our sensitivity analysis (App.~\ref{sec:robustness}) \citep[but see][]{barak2013sparseness,johnston2020nonlinear,fusi2016neurons}. In App.~\ref{sec:manifold-dimensionality}, we show that network activations follow the same spectral power law as cortical representations \citep{stringer2019high}. That is, they are low-dimensional enough to maintain differentiability (i.e. they are robust to input perturbations), while being high-dimensional enough to capture the data structure. This suggests a \textit{universal} coding strategy employed by biological and artifical neural networks alike. We believe that future analyses grounded in a quantified metric of interpretability may illuminate the computational function of these convergent neural coding strategies.

%% file: sections/6-acknowledgements.tex
\section*{Acknowledgements}    

We would like to thank Roland Zimmermann and Wieland Brendel for discussions, experiments with metrics and for sharing their psychophysics data.
Moreover, thanks to \citet{vinken2023neural} and \citet{higgins2021unsupervised} for publicly sharing their data and to Le Chang for helping with the extraction. 
We would also like to thank Katrin Franke and Andreas Tolias for discussions and feedback on the manuscript.
This work was supported by the U.S. Department of Energy, under DOE Contract No. DE-AC02-76SF00515, the SLAC National Accelerator Laboratory LDRD program, and the National Science Foundation under Grant 2313150.
Finally thanks to the complete Geometric Intelligence Lab at UCSB for providing feedback and support for this work.

%% file: sections/5-appendix.tex
\newpage

\section*{Appendix}
\appendix

\section{Task Explanation}
\label{sec:task_explanation}
Here is an illustration of the task used by \citet{zimmermann2023scale}.

\begin{figure}[H]
    \centering
    \includegraphics[width=0.6\textwidth]{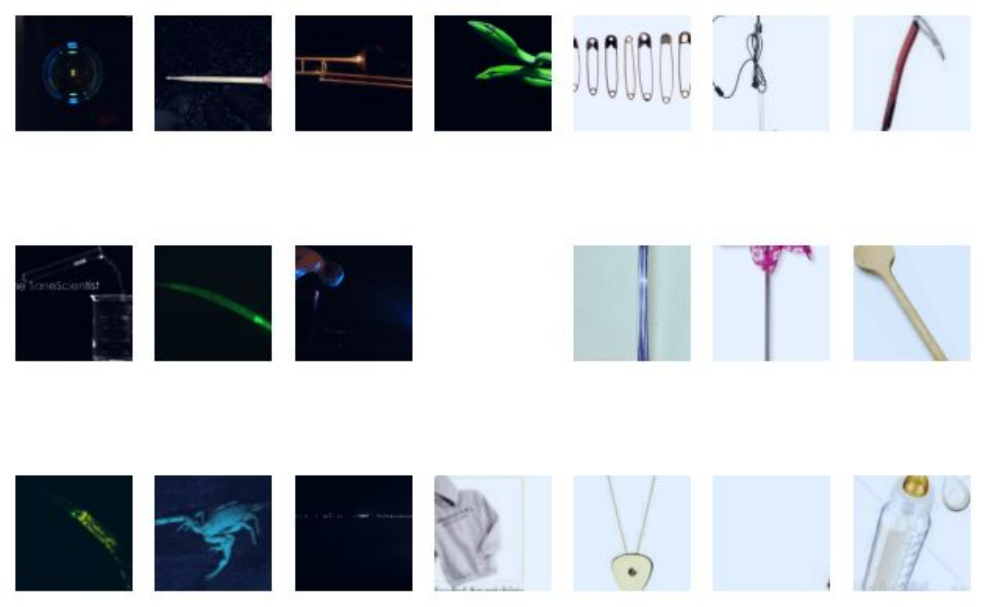}
    \caption{
    \textbf{Psychophysics Task.}
    Left images indicate positive reference ($MEI(y)$), right images indicate negative reference ($MEI(-y)$).
    The center images are the querries, the participant, here has to select the top image.
    }
    \label{fig:fig_task}
\end{figure}

\section{Comparison with a Sparse Autoencoder}
\label{sec:sparse_autoencoder}

We compare the K-Means approach for identifying interpretable directions in activation space to the shallow sparse autoencoder used in \cite{conjecture2022sparse_ae}.  The model is a single-layer  autoencoder trained with an L1 penalty on its hidden layer activation. 
We are training it for different numbers of hidden dimension and different values of sparsity regularisation.
All training is for $200$ epochs on the complete training set activations ($N=50,000$) with the Adam optimizer and a learning rate of $10^{-3}$.
We verified manually that these settings lead to convergence for all hyperparameter settings.

\begin{figure}[H]
    \centering
    \includegraphics[width=0.99\textwidth]{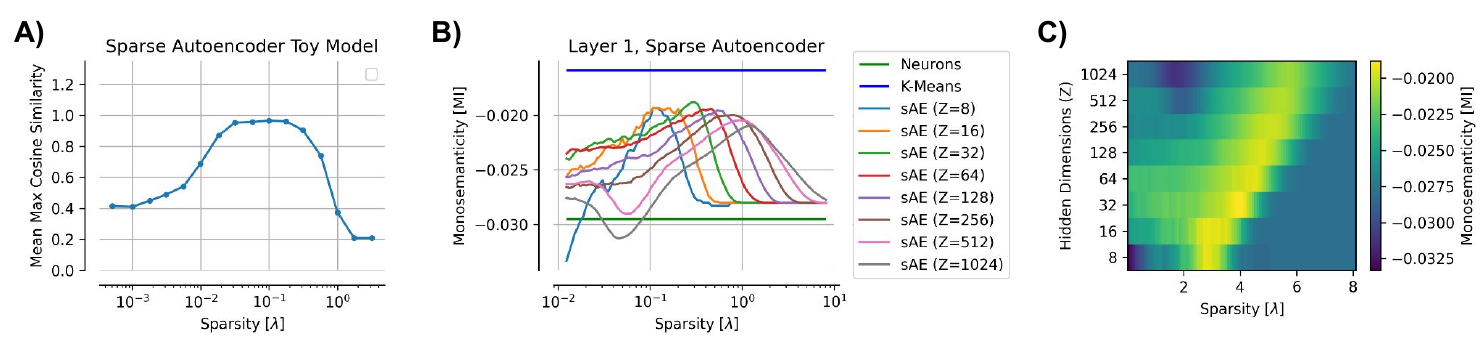}
    \caption{
    \textbf{Model Comparison with Sparse Autoencoder.}
    \textbf{A)} Sparse autoencoder results reproduced from \citep{conjecture2022sparse_ae}.
    \textbf{B), C)} We train the same sparse autoencoder model and measure the II index for different number of hidden dimensions ($Z$) and $L_1$ sparsity penalties ($\lambda$).
    }
    \label{fig:fig_sparse_ae}
\end{figure}

We see in  Fig.~\ref{fig:fig_sparse_ae} that the directions identified by the sparse autoencoder are more interpretable according to our metric than the original neuron basis. However, we find that the K-Means approach performs better than the sparse autoencoder.
In \cite{conjecture2022sparse_ae}, the authors assessed the relationship between source recovery and sparsity, using synthetic data containing known features. Here, we perform the same analysis, but with the II as a proxy for ground truth feature recovery. The functional relationship that we obtain between sparsity and II is remarkably similar Fig.~\ref{fig:fig_sparse_ae} a\&b, which suggests that the II may provide a good proxy for ground truth feature recovery.
This also aligns with our conclusions from Fig.~\ref{fig:higgins_dis_vs_psy}.

\section{Interpretable Neurons as Sparse Activations on the Data Manifold}
\label{sec:smt}

\begin{figure}[H]
    \centering
    \includegraphics[width=0.99\textwidth]{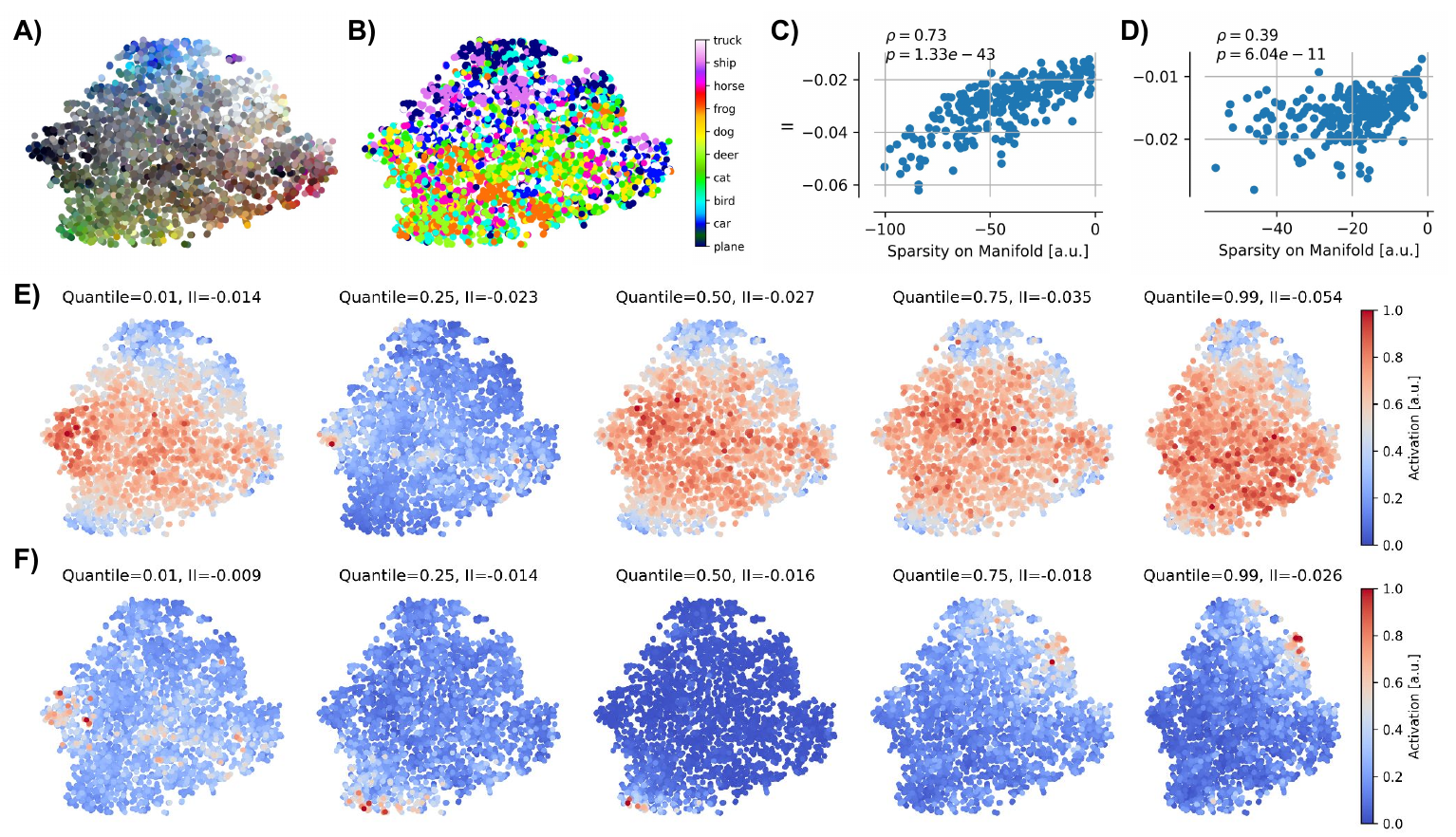}
    \caption{
    \textbf{Sparse Manifold Activations.}
    The natural image manifold (subset of MEIs) embedded in 2D with tSNE and coloured by average image colour \textbf{A)}, image label \textbf{B)}, activation of neurons \textbf{E)} and activation of features \textbf{F).}
    There is a correlation between sparsity on the manifold (average distance of most activating points) and the II for both neurons \textbf{C)} and features \textbf{D)}.
    }
    \label{fig:fig_manifold}
\end{figure}

One way of understanding interpretability is in terms of the distribution of a given neuron's activations over the image manifold. 
This concept and the following analysis are directly inspired by the \textit{sparse manifold transform} \citep{chen2018sparse}.
We take the top $M=5$ MEIs for both neurons ($N=256$ and K-Means ($K=256$) features and compute all pairwise image similarities using LPIPS. We then embed this distance matrix into a 2D space for visualization purposes using t-SNE \citep{van2008visualizing} (perplexity$=10$) (Fig. ~\ref{fig:fig_manifold}). Each point in the visualization corresponds to a different image, and is colored according to a different scheme in each subplot. In Fig. \ref{fig:fig_manifold}:A, the color of each point indicates the average color of the image. In Fig. \ref{fig:fig_manifold}:B, color indicates the image label. In Figs. \ref{fig:fig_manifold}:E and \ref{fig:fig_manifold}:F, color indicates the activation of a single neuron (E) or $K$-Means feature (F) over the dataset.
Activations for both neurons and $K$-Means features are computed as follows:
\begin{equation}
    f_i(x) = e^{-\frac{d'}{\tau}}, \quad d'=\frac{d-\text{avg}(d)}{\text{s.d.}(d)} d=\|x-\mu_i\|_2^2
\end{equation}
Where $\mu_i$ is the location of the cluster centroids for $K$-Means features and a one hot vector for neurons. Taking the z-score before the exponential ensures a fair comparison. 
Finally, the temperature $\tau=2$ is introduced for visualization purposes.\footnote{Lower $\tau$ would make Fig. \ref{fig:fig_manifold}:F look even sparser, and higher $\tau$ would make Fig. \ref{fig:fig_manifold}:E look even more uniform; using the same $\tau$ ensures a fair comparison.}

We see that colour (Fig. \ref{fig:fig_manifold}:A) is a major factor in determining the layout of the manifold, and although labels tend to cluster locally, image category plays a lesser role (Fig. \ref{fig:fig_manifold}:B).
Interestingly, we see that the features (Fig. \ref{fig:fig_manifold}:F) are much sparser on the manifold than the neural activations (Fig. \ref{fig:fig_manifold}:E). This suggests that more interpretable units are more sparsely active across the natural image manifold. Thus, we provide evidence (Fig. \ref{fig:fig_manifold}:C\&D) for a long-standing hypothesis in the neural coding literature \citep{chen2018sparse}.

\section{Robustness of Interpretable Features}
\label{sec:robustness}

\subsection{Sensitivity Analysis}

We investigate the sensitivity of the interpretable directions (directions), i.e., of the $K$-Means centroids. Specifically, we perturb each direction and quantify whether the perturbed direction is still interpretable. Our perturbation process is explained below. We interpolate from one $K$-Means centroid $\mu_a$ to another $\mu_b$ (and beyond to test extrapolation) and compute the II for these different directions in latent space. The intermediate directions are:
\begin{equation}
    v(\alpha) = \alpha \mu_a + (1 - \alpha) \mu_b, \quad \text{for $\alpha \in \mathbb{R}$}.
\end{equation}

\begin{figure}[H]
    \centering
    \includegraphics[width=0.99\textwidth]{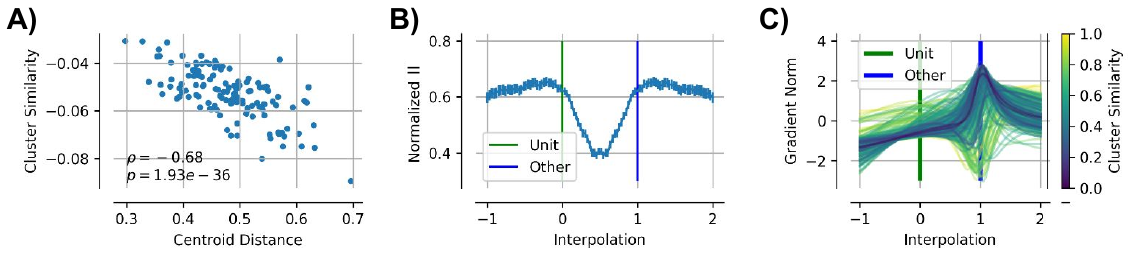}
    \caption{
    \textbf{Sensitivity Analysis.}
    \textbf{A)} Clusters that are further away from each other are have lower semantic similarity (measured as cross-II).
    \textbf{B)} Computing the II for interpolations (from one centroid to another) between and beyond all pairs of maximally separated clusters (determined with Hungarian algorithm).
    \textbf{C)} Sensitivity (i.e., average norm of the input gradient) for different interpolation points, coloured by the cluster similarity of start and end point in interpolation.
    }
    \label{fig:fig_sensitivity}
\end{figure}

We normalize each intermediate direction $v(\alpha)$ to maintain the same norm $1$.
For each direction $v$ along the interpolation path we compute the II index from the images that lie closest to that point in latent space:
\begin{equation}
    f_v(x) = - \|y(x) - v\|,
\end{equation}
where $x$ is an image (an input), $y$ is the function representing the first layer of the CNN, $y(x)$ is the feature associated to image $x$ in latent space, and $\|\|$ is the Euclidean norm in latent space. The results are shown in Figure~\ref{fig:fig_sensitivity} B). We observe that $\mu_a$ and $\mu_b$, corresponding to interpolation factors $\alpha = 0$ and $\alpha = 1$ have higher II and that the II strongly drops for interpolating directions $v(\alpha)$. This supports the idea that the direction extracted via $K$-Means are interpretable. Moreover, based on signal detection theory \citep{dayan2005theoretical}, we hypothesize that more meaningful directions are so in virtue of being highly \textit{selective} and less \textit{sensitive} to input perturbations, i.e., to image perturbations \cite{paiton2020selectivity}.
Thus, for each intermediate direction $v$, we also compute the norm of the input gradient:
\begin{equation}
    \| \nabla_{f_v}|_{x} \| = \| \nabla[- \|y(x) - v\|] \| = \| \nabla[- \|y(x) - \alpha \mu_a - (1 - \alpha) \mu_b\|] \|.
\end{equation}
For a fixed intermediate direction $v$, this value quantifies how much the response of a feature $y(x)$ representing image $x$ changes given perturbations on image $x$.

Figure~\ref{fig:fig_sensitivity} shows the average norm of the input gradient. We observe that the gradient's norm is generally much lower at a neuron's MEI ($\alpha=0$) vs. the MEI of a different unit ($\alpha=0$), unless they are very similar. We also find a weak but significant negative correlation (Spearman $\rho=-0.18$ $p<3.1 10^{-5}$ between the interpretability and the minimal gradient norm, as shown in Figure~\ref{fig:fig_sensitivity} D).
Together, these suggests that units which are more interpretable are also less sensitive to input perturbations at their preferred inputs.
Consequently, a hypothesis derived from these analyses: neurons in CNNs that are more interpretable are also more robust to adversarial or noise perturbations.

\subsection{Noise Robustness}

To test the hypothesis that more interpretable neurons are more noise robust, we add Gaussian noise to the inputs (standard deviation $\sigma \in [0, 0.1]$) and measure the sensitivity, i.e., the maximum absolute change in response compared to the clean image as proposed by \citep{guo2022adversarially}.
We then compare those scores to the II metric from the paper, as well as the logits from the \textit{in silico} psychophysics task.
We find a weak but significant relationship in both cases.
This supports the hypothesis that more interpretable neurons are more robust to white noise input perturbations.

\begin{figure}[H]
    \centering
    \includegraphics[width=0.99\textwidth]{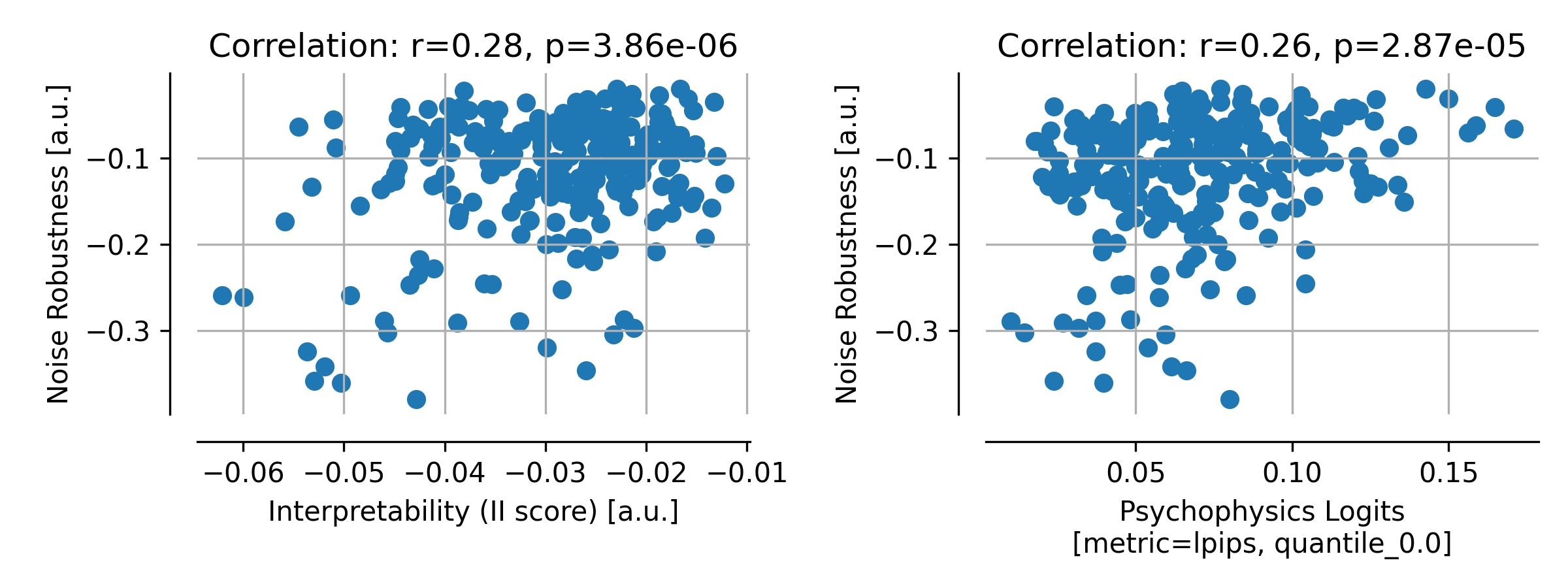}
    \caption{
    \textbf{Noise Robustness of Neurons.}
    Left, noise robustness \citep{guo2022adversarially} of neurons, plotted as a function of their II scores (see paper); linear correlation indicated above.
    Right, same but as a function of psychophysics logits (for largest quantile and LPIPS metric, see main text).
    }
    \label{fig:fig_noise_robustness}
\end{figure}

\section{Monosemanticity and the Privileged Basis Hypothesis}
\label{sec:privileged-basis}

Why should we see individual neurons learning meaningful representations at all? Recent research in the mechanistic interpretability literature has suggested that there exists \textit{privileged bases} in neural networks, corresponding to neurons, emerging from nonlinearities such as ReLU that operate per neuron. The intuition behind the privileged basis needs further explanation.

\begin{figure}[H]
    \centering
    \includegraphics[width=0.8\textwidth]{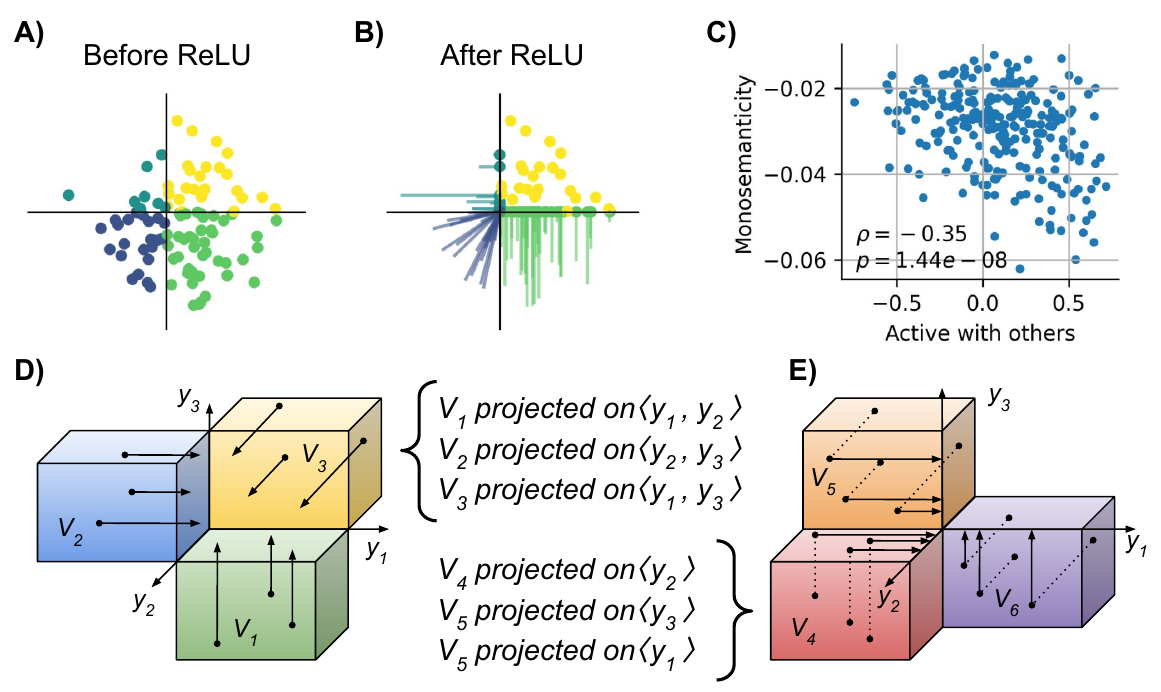}
    \caption{
    \textbf{Privileged Basis Hypothesis.}
    Activations before \textbf{A)} and \textbf{B)} after ReLU nonlinearity. In this scenario, the first neuron is more active as it has to represent the large number of (green) points in the bottom right quadrant, while the second neuron only needs to represent few (blue) points in the top left quadrant.
    \textbf{C)} Significant negative correlation between neurons that are more active together with others (measured as the correlation between the neuron's activity and the population, excluding the neuron, average) and the interpretability index (II).
    \textbf{D} In a three dimensional space, the positive only quadrant remains untouched by ReLUs, the negative only quadrant gets mapped to 0, three quadrants ($V_1, V_2, V_3$, \textbf{D}) are projected onto two dimensional subspaces, and three quadrants ($V_4, V_5, V_6$, \textbf{D}) are projected onto one dimensional subspaces.
    Thus, e.g. neuron $y_1$ has to represent (together with $y_2$) all of $V_1$, and it also has to represent (completely on its own) all of $V_6$.
    }
    \label{fig:fig_privileged_basis}
\end{figure}

For $K$ neurons, there are $2^K$ quadrants as shown in Figure~\ref{fig:fig_privileged_basis} A) for $K=2$. We consider what happens for a feature vector in each of these quadrant after the application of ReLU.
Of those, $1$ (all positive) stays untouched, as shown in yellow in Figure~\ref{fig:fig_privileged_basis} B. Another $1$ quadrant (all negative) becomes zero: shown in purple. Next, $K$ quadrants get represented by $1$ neuron (i.e. $K$ dimensions are collapsed to $1$) shown in blue and green; $K-1$ get represented by $2$ neurons (i.e. $K$ dimensions are collapsed to $1$), etc. In other words, each neuron participates in encoding points in $K!$ quadrants of compressed dimensionality. Now, we ask: which of these neurons should be more interpretable?

Consider data points unequally distributed into the different quadrants, with one quadrant (bottom right in Figure~\ref{fig:fig_privileged_basis} A)) containing more points than another (top left in Figure~\ref{fig:fig_privileged_basis} A)). Neurons in charge of representing the features from a quadrant with many point is more active. A resulting hypothesis is that neurons which are more active when others are inactive, i.e., which are active alone, are more interpretable | and form the priviledged basis.

Figure~\ref{fig:fig_privileged_basis} C) uses our our interpretability index (II) to confirm this hypothesis by showing a negative correlation between a co-activation measure (``Active with Others'') and II. The co-activation measure is defined as the correlation between each neuron's response and the average population response.

\section{Dimensionality of the Neural Activation Manifold}

\label{sec:manifold-dimensionality}

\begin{figure}[H]
    \centering
    \includegraphics[width=14cm]{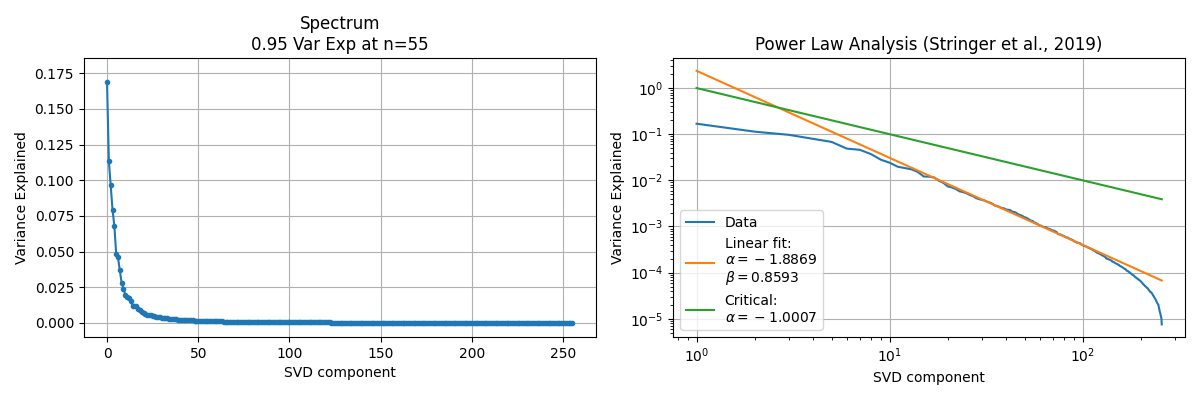}
    \caption{
    \textbf{High Dimensional Smooth Activation Manifold.}
    Same analysis as in \cite{stringer2019high}, showing that activations in CNN feature space are high dimensional within the constraints of remaining differentiable.
    A spectrum that decays slower than the critical value (green line in right plot, \cite{stringer2019high}), would be non-differentiable and, therefore, highly non-robust.
    Remarkably, this spectral behaviour is the same as observed across many cortical areas.
    }
    \label{fig:layer1_svd}
\end{figure}

\newpage
\section{Additional Comparisons to Psychophysics Data}
Here we look at some of the other exemplary models studied by \citet{zimmermann2023scale}.

\begin{figure}[H]
    \centering
    \includegraphics[width=0.99\textwidth]{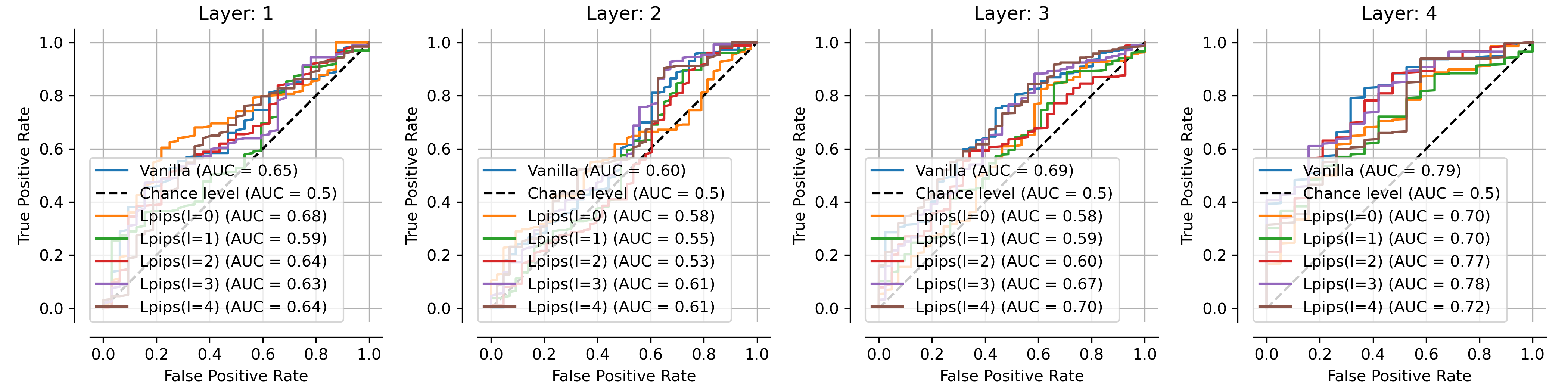}
    \includegraphics[width=0.99\textwidth]{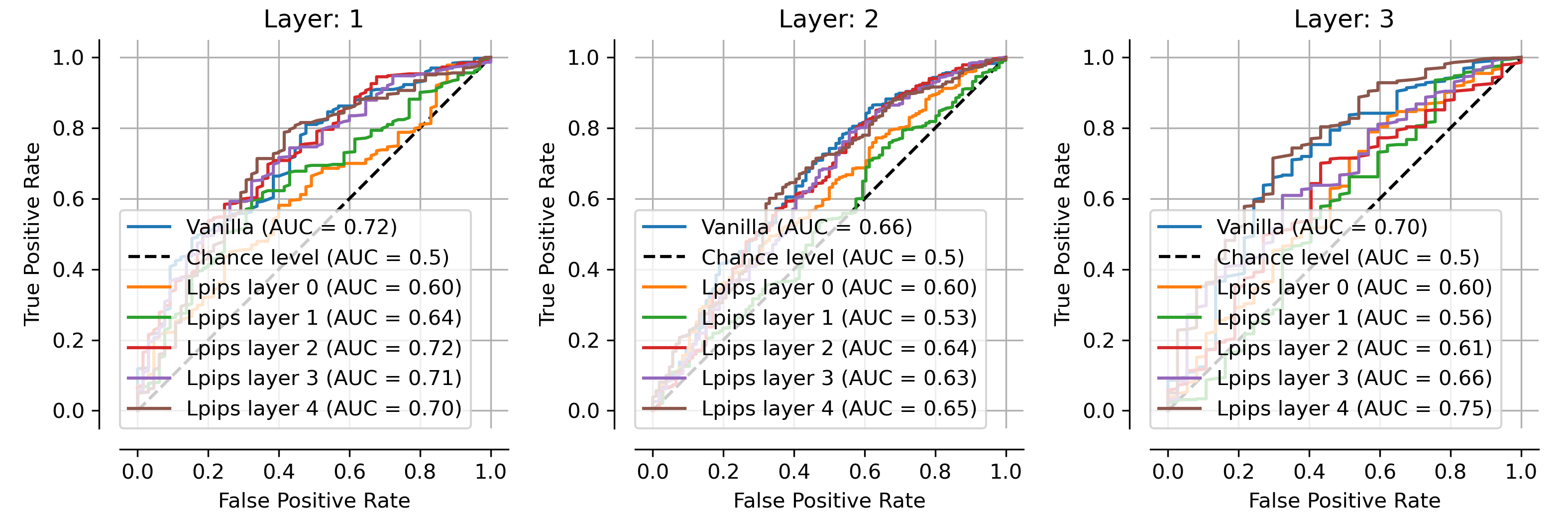}
    \caption{
    \textbf{Psychophysics Match for Other Models.}
    Data from \cite{zimmermann2023scale}. 
    Top, Clip ResNet50; Bottom, GoogLeNet.
    Left to Right: Agreement between human and in-silico psychophysics on the predictability of the outputs of different layers in the network. Human and model agree on what makes a feature predictability for the early layers. For these layers, the proposed interpretability metric is a valid representation of the human's perception of interpretability. 
    AUC: Area Under the Curve.
    }
    \label{fig:fig_human_psychphysics_other}
\end{figure}

\newpage
\section{Disentangling Model and Data}
\label{sec:untrained_model}
Like Bricken et al.\ \citep{bricken2022monosemanticity}, we run our analyses again on an untrained model to assess how much of the interpretability results we obtained are due to training (model) versus derived from properties of the data.
Interestingly, we find that, indeed, the gap in interpretability between neurons and activity clusters is already present at initialization.
This confirms prior experiments that showed that CNNs are, even untrained, already very useful representations of image data \citep{frankle2020training}.
However, we note a clear training effect: the semantic level of interpretability shifts over the course of training.
That is, while the untrained model neurons and features are easily predicted by low level image properties, such as color, the trained model is better predicted by high level semantics, such as label information.

\begin{figure}[H]
    \centering
    \includegraphics[width=0.99\textwidth]{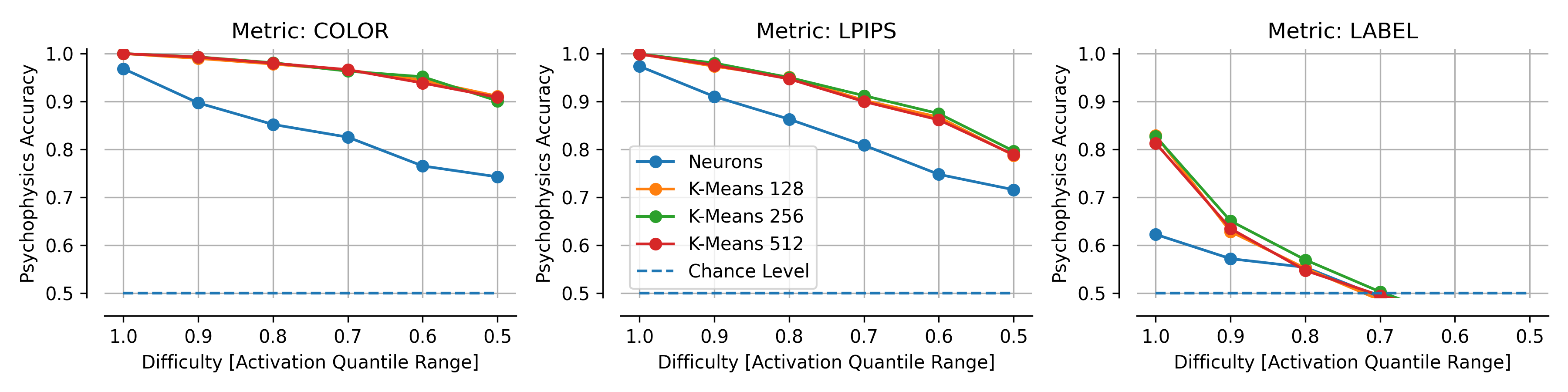}
    \includegraphics[width=0.99\textwidth]{figures/fig_psycho.png}
    \caption{
    \textbf{Untrained Psychophysics.}
    Top, untrained model; Bottom, trained model, same as Fig.\ref{fig:fig_psycho}.
    The interpretability gap is already apparent at initialisation, however, there is a clear shift in semantics from low level (color) at initialisation, to high level (label) after training.
    }
    \label{fig:fig_psycho_untrained}
\end{figure}